\documentclass{article}

    \PassOptionsToPackage{numbers, compress}{natbib}

\usepackage[final,main]{neurips_2026}

\usepackage[utf8]{inputenc} 
\usepackage[T1]{fontenc}    
\usepackage{hyperref}       
\usepackage{url}            
\usepackage{booktabs}       
\usepackage{amsfonts}       
\usepackage{nicefrac}       
\usepackage{microtype}      
\usepackage{xcolor}         

\usepackage{graphicx} 
\usepackage{microtype}
\usepackage{colortbl}
\usepackage{hyperref}
\usepackage{url}
\usepackage{booktabs}
\usepackage{xcolor}
\usepackage{algorithm}
\usepackage{float}
\hypersetup{hidelinks}
\usepackage{amsmath}
\usepackage[noend]{algpseudocode}  
\usepackage{tcolorbox}
\usepackage{multicol}
\usepackage{multirow}
\usepackage{adjustbox}
\usepackage{amssymb} 
\usepackage{subcaption}
\usepackage{enumitem} 
\definecolor{pastelblue}{RGB}{173,216,230}
\definecolor{pastelpink}{RGB}{255,182,193}
\definecolor{pastelgreen}{RGB}{152,251,152}
\usepackage{threeparttable}
\usepackage{booktabs}
\usepackage{multirow}
\usepackage{wrapfig}
\usepackage{enumitem}

\usepackage{pifont}
\newcommand{\cmark}{\ding{51}}
\newcommand{\xmark}{\ding{55}}

\title{BiMoGen: Bidirectional Motion-Text Generation via Unified Masked Discrete Diffusion} 

\author{%
  \textbf{Wanjiang Weng}$^{1,2,*}$ \quad
  \textbf{Yongliang Wu}$^{1,2,*}$ \quad
  \textbf{Xiaofeng Tan}$^{1,2}$ \\
  \textbf{Xingyu Zhu}$^{3}$ \quad
  \textbf{Wenbo Zhu}$^{4}$ \quad
  \textbf{Hongsong Wang}$^{1,2,\dagger}$ \\[0.em]
  \parbox{\dimexpr\textwidth-2\tabcolsep\relax}{\centering\normalfont\small
    $^{1}$Department of Computer Science and Engineering, Southeast University, Nanjing, China\\[0pt]
    $^{2}$Key Laboratory of New Generation Artificial Intelligence Technology and Its Interdisciplinary \\ Applications (Southeast University), Ministry of Education, Nanjing, China\\[0pt]
    $^{3}$National University of Singapore\\[0pt]
    $^{4}$HeyGen
  }
}
\begin{document}

\maketitle

\begingroup
  \renewcommand{\thefootnote}{\fnsymbol{footnote}}
  \footnotetext[1]{Equal contribution.}
  \footnotetext[2]{Corresponding author.}
\endgroup

\vspace{-10pt}
\begin{abstract}
Text-to-motion generation and motion-to-text captioning are two fundamental tasks in human motion modeling, both grounded in the same underlying motion-text correspondence. Existing unified approaches mostly rely on autoregressive modeling, which imposes a fixed generation order and is therefore poorly suited to the bidirectional dependencies between language and motion, allowing early prediction errors to persist as fixed context and degrade both temporal coherence and cross-modal consistency. Masked discrete diffusion, which models sequences through iterative bidirectional prediction, allows prediction from bidirectional context without a fixed generation order. We therefore propose \textbf{BiMoGen} (\textbf{Bi}directional \textbf{Mo}tion-text \textbf{Gen}eration), a unified masked discrete diffusion framework for bidirectional motion-text modeling. To stabilize training, we design Decoupled Uni- and Cross-Modal Training, in which masked pretraining first establishes cross-modal correspondence on paired motion-text sequences, after which supervised fine-tuning specializes the model for bidirectional generation. Masked diffusion nonetheless faces a train-test mismatch, as the model is trained on clean ground-truth context yet encounters self-generated and potentially erroneous context at inference, with errors committed under heavily masked states propagating through subsequent steps. We further introduce Generation-Aware Self-Correction that exposes the model to its own predictions during training and uses correction passes at early sampling steps for further refinement of previously unmasked tokens. Extensive experiments on HumanML3D and KIT-ML demonstrate competitive performance on both tasks, validating the effectiveness of the proposed two-stage training and self-correction designs. The project page is available at \url{https://wengwanjiang.github.io/BiMoGen-Page}.

\end{abstract}
\section{Introduction}
\label{sec:intro}
Text-to-motion generation (T2M) and motion-to-text captioning (M2T) are two closely related tasks that both rely on understanding the correspondence between human motion and natural language~\cite{realign,wang2025foundation,lamp,tan2025sopo,eztune}. Recent work has therefore pursued unified frameworks that address both tasks within a single model, so that both conditional directions can use the same backbone and motion-text token space. Many existing approaches build on autoregressive modeling~\cite{mgpt,mgpt3,mgpt2,zheng2026next}. Token-based approaches discretize motion, concatenate motion and text tokens, and train a Transformer through next-token prediction~\cite{mgpt,mgpt2,motionllama}. Yet the underlying causal backbone imposes a fixed left-to-right generation order and provides no mechanism to revise early prediction errors. Once a token is incorrectly produced, it remains in the prefix conditioning all subsequent steps, and the resulting errors propagate through the sequence and ultimately degrade temporal coherence and motion-text consistency.

Masked discrete diffusion offers a more principled alternative~\cite{wu2023ar,sedd, d3pm, llada,maskgit,simplediffusion,arriola2025block}. By iteratively predicting masked tokens from bidirectional context, it naturally captures the bidirectional dependencies between text and motion without imposing a fixed left-to-right generation order. However, applying masked diffusion to unified motion-text modeling raises two fundamental challenges. First, training a single model for both T2M and M2T entangles cross-modal correspondence learning with conditional generation, forcing the model to produce target tokens before reliable motion-text alignment has emerged and destabilizing optimization. Second, the model is trained on corrupted ground-truth context yet must condition on its own earlier predictions at inference, and this train-test misalignment becomes most severe under the high mask ratios encountered in early sampling steps, where committed errors persist and cascade into later predictions.

\begin{figure}[t]
    \centering
    \includegraphics[width=\linewidth]{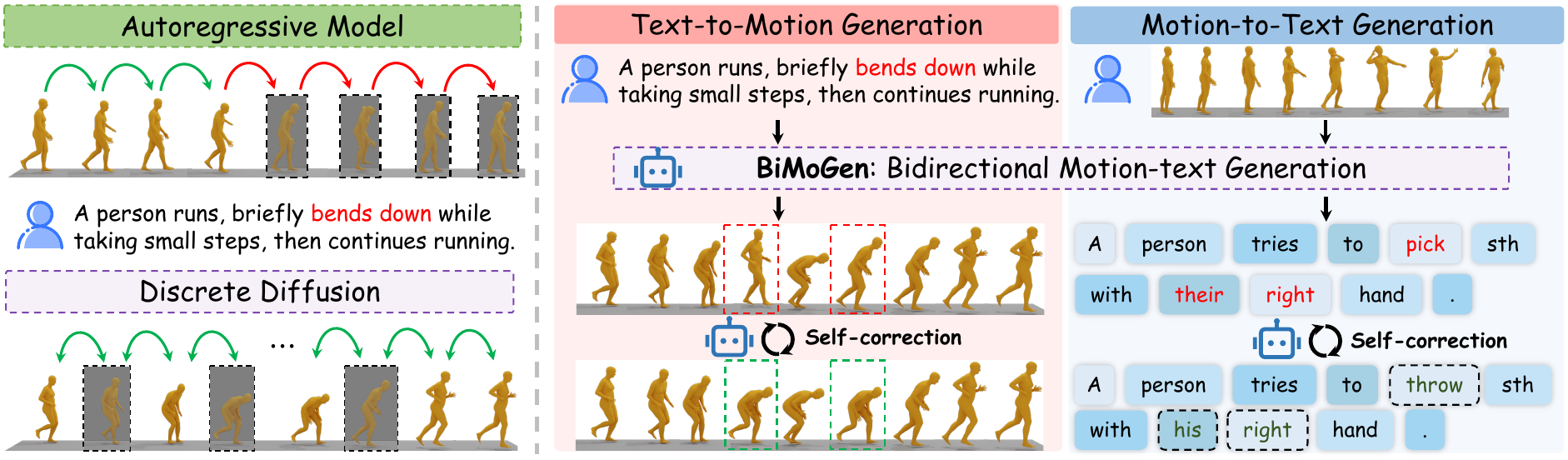}
    \caption{Overview of the BiMoGen framework. BiMoGen unifies text-to-motion generation and motion-to-text captioning within a single masked discrete diffusion model, and employs a self-correction mechanism that revises unreliable predictions at early sampling steps.}
    \label{fig:overview}
    
\end{figure}

To address these challenges, we propose \textbf{BiMoGen} (\textbf{Bi}directional \textbf{Mo}tion-text \textbf{Gen}eration), a unified masked discrete diffusion framework for bidirectional motion-text modeling. As shown in Figure~\ref{fig:overview}, BiMoGen handles both T2M and M2T within a single model through masked token prediction, naturally accommodating the bidirectional dependencies between the two modalities. Recognizing that cross-modal alignment and conditional generation impose fundamentally different learning objectives, we introduce a two-stage strategy named Decoupled Uni- and Cross-Modal Training. The first stage performs masked pretraining on paired motion-text sequences to establish robust cross-modal correspondence, while the second stage applies supervised fine-tuning that specializes the model for bidirectional generation. To further close the gap between training and inference, we propose Generation-Aware Self-Correction that explicitly aligns the two regimes. During training, the model learns to recover ground-truth tokens from its own predicted context, while at inference correction passes at early sampling steps provide further refinement of previously unmasked tokens. Extensive experiments on HumanML3D~\cite{humanml3d} and KIT-ML~\cite{plappert2016kit} demonstrate that BiMoGen achieves competitive performance on both T2M generation and M2T captioning.

Our contributions are summarized as follows:
\begin{itemize}[leftmargin=2em,topsep=0pt]
\setlength{\itemsep}{0pt}
\setlength{\parsep}{0pt}
\setlength{\parskip}{0pt}
    \item We propose BiMoGen, a unified masked discrete diffusion framework that formulates T2M generation and M2T captioning under a shared bidirectional architecture.

    \item We introduce Generation-Aware Self-Correction, applied during training to mitigate the train-test misalignment inherent in masked diffusion and during sampling for further refinement.

    \item We design Decoupled Uni- and Cross-Modal Training, which comprises masked pretraining and supervised fine-tuning stages to disentangle cross-modal alignment from conditional generation.

\end{itemize}
\section{Related Works}
\noindent \textbf{Text-to-Motion Generation.} 
Generating 3D human motion from natural language has emerged as a fundamental task with applications in animation, virtual humans, and embodied agents~\cite{wu2025finemotion,kinmo2025,realign,usdrl,kuang2025zero,kuang2026genprior, motionclip, motionlcm,fan2025go}. Continuous-diffusion approaches model motion either in raw coordinate space~\cite{meng2025absolute,meng2024rethinking,ruiz2025mixermdm}, as in MDM~\cite{mdm}, or in a learned latent space for efficiency, as in MLD~\cite{mld}. A parallel line of work discretizes motion into tokens via VQ-VAE~\cite{vqvae} and models them with sequence models~\cite{pinyoanuntapong2024bamm,kong2023priority}. T2M-GPT~\cite{t2mgpt} adopts a GPT-style autoregressive decoder, whereas MoMask~\cite{momask} introduces residual quantization with masked generative modeling. BAMM~\cite{pinyoanuntapong2024bamm} combines autoregressive generation with bidirectional refinement, while BiPO~\cite{bipo} uses partial occlusion and selective token regeneration. Their bidirectional prediction is trained on ground-truth context, whereas GASC explicitly trains recovery from self-generated context. Despite the diversity of generative paradigms, these methods are tailored to unidirectional text-to-motion synthesis and provide neither motion-to-text captioning nor a shared formulation that exploits the bidirectional dependencies between language and motion.

\noindent \textbf{Unified Motion-Text Modeling.} 
Beyond unidirectional synthesis, recent work studies unified models that handle motion generation and captioning within a shared framework~\cite{mgmotionllm,mgpt2,zxy1,dft,zhang2026towards,unimo,tan2026perceptualentropy,chen2025language,motionllama,yu2025remogpt,wang2026unimotion}. TM2T~\cite{tm2t} introduces motion tokens and trains separate translators for the two directions. MotionGPT~\cite{mgpt} treats motion as a foreign language and unifies T2M and M2T under an autoregressive Transformer, while MotionGPT3~\cite{mgpt3} extends this paradigm with a lightweight diffusion head to improve motion fidelity. However, the underlying causal backbone still imposes a fixed left-to-right generation order, which is misaligned with the inherently bidirectional dependencies between text and motion. Unimotion~\cite{li2024unimotion} jointly diffuses motion and frame-level text representations for motion synthesis and understanding. A concurrent work, DiMo~\cite{dimo}, also adopts masked discrete diffusion for bidirectional motion-text modeling, using an RVQ predictor and reinforcement-learning fine-tuning to enhance motion fidelity and cross-modal alignment. BiMoGen instead focuses on the train-test misalignment and error accumulation of masked diffusion. By introducing generation-aware self-correction during both training and sampling, BiMoGen enables the model to revise unreliable predictions while retaining a simpler single-Transformer design over a single-layer VQ-VAE.

\noindent \textbf{Masked Discrete Diffusion Models.}
Masked discrete diffusion generates discrete sequences by iteratively predicting masked tokens with a bidirectional Transformer, offering a nonautoregressive alternative to GPT style decoding~\cite{cdcd,huunified,diffusionbert,gong2024scaling,genie,wang2025fudoki,lavida,wu2023ar}. D3PM~\cite{d3pm} formalizes diffusion over discrete state spaces, MaskGIT~\cite{maskgit} introduces confidence guided parallel decoding for image synthesis, and SEDD~\cite{sedd} and LLaDA~\cite{llada} extend related approaches to language modeling. GIDD~\cite{gidd} explores corruption with both masking and uniform noise alongside correction of generated tokens, while GASC retains mask only corruption and supervises predictions made from generated context. For bidirectional modeling of motion and text, the large difference in sequence length can bias unified prediction toward motion reconstruction, while errors in generated motion tokens can accumulate during iterative refinement. BiMoGen addresses these challenges through decoupled cross modal pretraining, conditional fine tuning, and generation-aware self correction.

\section{Method}
In this section, we present BiMoGen, a unified masked discrete diffusion framework for bidirectional motion-text modeling. We first formulate the unified bidirectional motion-text generation task (Sec.~\ref{sec:formulation}), then introduce the decoupled training procedure (Sec.~\ref{sec:training}), and finally present generation-aware self-correction for training and sampling (Sec.~\ref{sec:inference}).

\subsection{Unified Masked Discrete Diffusion Model}\label{sec:formulation}

\textbf{Discrete Representation.} We employ a pretrained VQ-VAE~\cite{vqvae,mgpt} as the motion tokenizer, whose encoder $\mathcal{E}$ discretizes a motion sequence $\mathbf{m} = (m_1, \ldots, m_F)$ of $F$ frames into a sequence of $N_m = F/r$ motion tokens $\mathbf{x}^m = (x^m_1, \ldots, x^m_{N_m})$, where $r$ denotes the temporal downsampling rate, and whose decoder $\mathcal{D}$ maps motion tokens back to a continuous motion sequence at inference time. Let $\mathbf{x}^c = (x^c_1, \ldots, x^c_{N_c})$ denote the corresponding text tokens of length $N_c$. The motion vocabulary is appended to the text vocabulary so that both modalities share a unified token space~\cite{mgpt}. The two modalities are concatenated into a single sequence $\mathbf{x}_0 = [\mathbf{x}^c, \mathbf{x}^m]$ of length $N = N_c + N_m$, which serves as the unified input for masked discrete diffusion.

\textbf{Forward Process and Training Objective.} We follow the masked discrete diffusion formulation~\cite{llada}. For a timestep $t$ sampled uniformly from $(0, 1]$, each token in $\mathbf{x}_0$ is independently replaced by a mask token $\texttt{[M]}$ with probability $t$, yielding a corrupted sequence $\mathbf{x}_t$. A bidirectional Transformer $p_\theta(\cdot \mid \mathbf{x}_t)$ outputs a distribution over clean tokens at every position and is optimized by
\begin{equation}\label{eq:lmdm}
    \mathcal{L}_{\mathrm{MDM}}(\theta) = -\mathbb{E}_{t, \mathbf{x}_0, \mathbf{x}_t}\left[\frac{1}{t} \sum_{i=1}^{N} \mathbf{1}[x^i_t = \texttt{[M]}] \log p_\theta(x^i_0 \mid \mathbf{x}_t)\right],
\end{equation}
where the indicator $\mathbf{1}[\cdot]$ restricts supervision to masked positions and the factor $1/t$ arises from the variational bound and reweights contributions across different mask ratios. The corresponding reverse process iteratively predicts masked tokens conditioned on currently visible ones, which we instantiate at sampling time in Sec.~\ref{sec:inference}.

\begin{figure}[t]
    \centering
    \includegraphics[width=\linewidth]{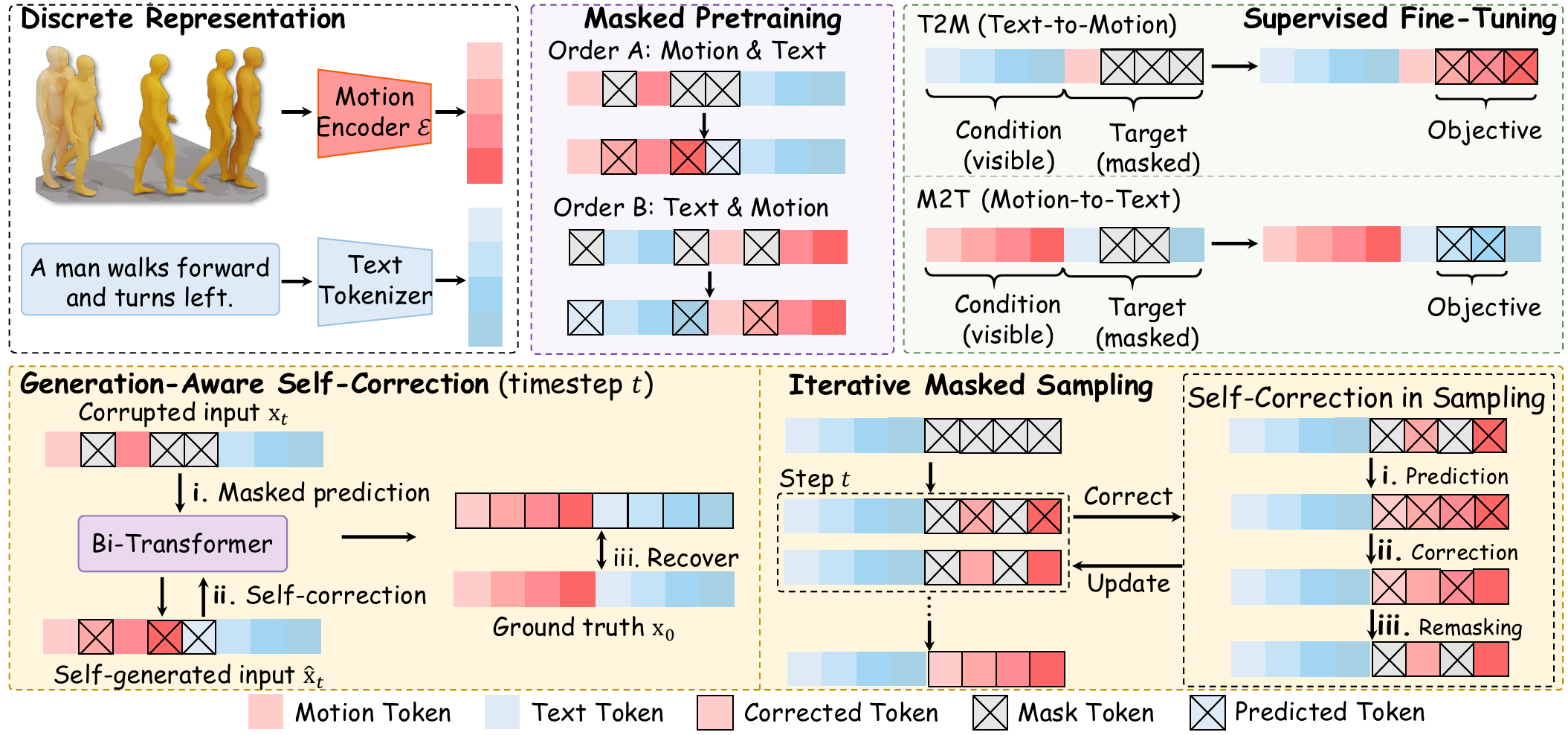}
    \caption{Overview of the BiMoGen framework. \textbf{Top:} motion and text are represented as discrete tokens in a unified vocabulary, and the model is trained in two-stages, masked pretraining over symmetrically corrupted motion-text sequences and supervised fine-tuning for T2M and M2T generation. \textbf{Bottom:} a self-correction mechanism is applied at both training and sampling, where the model takes its own predictions as input and learns to recover the ground-truth, and analogously revises committed tokens during iterative masked decoding.}
    \label{fig:framework}
\end{figure}

\subsection{Decoupled Uni- and Cross-Modal Training}\label{sec:training}
Bidirectional motion-text generation ultimately requires the model to produce one modality conditioned on the other, but conditional generation can only succeed once a reliable cross-modal correspondence has been established. Optimizing both objectives jointly forces the model to generate target tokens before alignment has emerged, which destabilizes training. We therefore introduce Decoupled Uni- and Cross-Modal Training (DUCMT): masked pretraining first learns motion-text correspondence over unified sequences, and supervised fine-tuning then specializes the model for the T2M and M2T directions.

\textbf{Masked Pretraining for Motion-Text Correspondence.} The first stage jointly models text and motion under the masked prediction objective in Eq.~\eqref{eq:lmdm}, with the loss denoted $\mathcal{L}_{\mathrm{PT}}$. Each paired sequence is formed by concatenating $\mathbf{x}^c$ and $\mathbf{x}^m$ in either order with equal probability, which prevents the model from relying on a fixed modality order as a positional prior. Both modalities are corrupted by the same forward process, and the model is supervised on all masked positions regardless of modality. This symmetric corruption provides direct token-level supervision to both sides and encourages the Transformer to learn motion-text correspondence.

\textbf{Supervised Fine-tuning for Bidirectional Generation.} The second stage adapts the pretrained backbone to conditional generation by fine-tuning on pairs $(\mathbf{x}^{\mathrm{cond}}, \mathbf{x}^{\mathrm{tgt}})$. For T2M, the condition is text $\mathbf{x}^c$ and the target is the motion $\mathbf{x}^m$. For M2T, the assignment is reversed. The condition is kept visible while only the target is corrupted by the forward process, and the masked prediction objective is computed only at masked target positions. The supervised fine-tuning loss is formulated as
\begin{equation}
    \mathcal{L}_{\mathrm{SFT}}(\theta) = -\mathbb{E}_{t, \mathbf{x}_0, \mathbf{x}^{\mathrm{tgt}}_t}\left[\frac{1}{t} \sum_{i=1}^{N_{\mathrm{tgt}}} \mathbf{1}[x^{\mathrm{tgt},i}_t = \texttt{[M]}] \log p_\theta\!\left(x^{\mathrm{tgt},i}_0 \mid \mathbf{x}^{\mathrm{cond}}, \mathbf{x}^{\mathrm{tgt}}_t\right)\right],
\end{equation}
where $N_{\mathrm{tgt}}$ denotes the length of $\mathbf{x}^{\mathrm{tgt}}$. To support classifier-free guidance at sampling time, we further replace the entire condition with $\texttt{[M]}$ tokens at a $10\%$ dropout rate so that the model jointly learns the conditional and unconditional distributions~\cite{cfg}.

\textbf{Auxiliary Motion-only Supervision.} Beyond the two cross-modal directions, we include a motion-to-motion (M2M) objective, where motion tokens at randomly sampled positions serve as the condition and the rest as the target. Text descriptions capture selected aspects of a motion sequence, whereas M2M provides dense intra-modal supervision to strengthen the motion prior of the unified model.

\subsection{Generation-Aware Self-Correction}\label{sec:selfcorrection}
Both training stages above supervise $p_\theta$ on inputs whose visible positions hold ground-truth tokens, whereas at sampling time visible target positions are filled by tokens that $p_\theta$ predicted in earlier steps. This train-test misalignment between training on ground-truth context and inference on self-generated context is a core limitation of masked diffusion training. Once a token is unmasked during sampling, it remains as fixed context for all subsequent steps, and an error made under high mask ratios propagates rather than gets corrected. We address this mismatch with Generation-Aware Self-Correction (GASC), which trains the model to recover ground-truth tokens from its own predictions and further refines previously unmasked tokens during sampling.

\textbf{Training with Self-Correction.} 
Let $\mathcal{G}$ denote the generation region of the current stage, covering the entire sequence in pretraining and only the target in supervised fine-tuning. Given a corrupted input $\mathbf{x}_t$, the model first performs the standard masked prediction pass. We then construct a self-generated input $\tilde{\mathbf{x}}$ by replacing tokens in $\mathcal{G}$ with the model's detached argmax predictions, while tokens outside $\mathcal{G}$ remain unchanged
\begin{equation}
    \tilde{x}^i = \mathrm{sg}\!\left(\arg\max_v\, p_\theta(v \mid \mathbf{x}_t)_i\right), \quad i \in \mathcal{G},
\end{equation}
where $\mathrm{sg}(\cdot)$ stops gradients through the discrete prediction. The same Transformer is then applied to $\tilde{\mathbf{x}}$ and trained to recover the ground-truth tokens over the entire generation region
\begin{equation}
    \mathcal{L}_{\mathrm{SC}}(\theta) = -\mathbb{E}_{t, \mathbf{x}_0, \tilde{\mathbf{x}}}\left[\frac{1}{t} \sum_{i \in \mathcal{G}} \log p_\theta(x^i_0 \mid \tilde{\mathbf{x}})\right].
\end{equation}
Unlike the standard masked diffusion loss, this correction loss is applied to all positions in $\mathcal{G}$, since committed tokens during sampling may also be incorrect and should remain revisable. The full objective for each training stage is
\begin{equation}
    \mathcal{L}(\theta) = \mathcal{L}_{\mathrm{stage}}(\theta) + \lambda\, \mathcal{L}_{\mathrm{SC}}(\theta), \quad \mathrm{stage} \in \{\mathrm{PT},\, \mathrm{SFT}\},
\end{equation}
where $\lambda$ controls the strength of self-correction supervision.

\textbf{Sampling with Self-Correction.}
\label{sec:inference}
At inference time, BiMoGen generates the target sequence by iteratively refining a fully masked target. Given a condition $\mathbf{x}^{\mathrm{cond}}$, we initialize $\mathbf{x}^{\mathrm{tgt}}$ as a sequence of $\texttt{[M]}$ tokens. For T2M, the target motion length is given. For M2T, we generate up to a maximum text length and truncate the output at the first $\texttt{[EOS]}$ token.

Sampling proceeds for $T$ uniformly spaced timesteps. At each timestep, $p_\theta$ predicts a distribution over clean tokens at all masked positions. We compute the confidence of each masked position as the maximum predicted probability and progressively unmask the most confident positions, while leaving the remaining positions masked for subsequent steps. We further apply classifier-free guidance~\cite{cfg,mld} during sampling, where the conditional and unconditional predictions are linearly combined as
\begin{equation}
    p^w_\theta\!\left(\cdot \mid \mathbf{x}^{\mathrm{cond}}, \mathbf{x}^{\mathrm{tgt}}\right) = w\, p_\theta\!\left(\cdot \mid \mathbf{x}^{\mathrm{cond}}, \mathbf{x}^{\mathrm{tgt}}\right) + (1 - w)\, p_\theta\!\left(\cdot \mid \emptyset, \mathbf{x}^{\mathrm{tgt}}\right),
\end{equation}
where $w$ is the guidance scale and $w > 1$ strengthens the conditioning effect. The unconditional case $\emptyset$ corresponds to replacing the condition with $\texttt{[M]}$, matching the dropout used during fine-tuning.

Iterative sampling uses unmasked tokens as context for subsequent steps, so errors made under heavily masked states can affect later predictions. We address this by invoking self-correction at a set of selected sampling steps $\mathcal{R} \subseteq \{1, \ldots, T\}$. When triggered, the current partially masked target is first predicted into a complete target candidate $\tilde{\mathbf{x}}^{\mathrm{tgt}}$, which is then fed back to the model for correction. The corrected prediction is written back only to positions that have already been unmasked
\begin{equation}
    x^{\mathrm{tgt}, i} \leftarrow \arg\max_v\, p_\theta\!\left(v \mid \mathbf{x}^{\mathrm{cond}}, \tilde{\mathbf{x}}^{\mathrm{tgt}}\right)_i, \quad \text{for } x^{\mathrm{tgt}, i} \neq \texttt{[M]}.
\end{equation}

The remaining masked positions are kept as $\texttt{[M]}$ and deferred to subsequent sampling steps. This correction step allows tokens committed in earlier iterations to be revised as the context becomes clearer without prematurely filling unresolved positions. After iterative decoding, the predicted motion tokens are decoded back to a continuous motion sequence by the VQ-VAE decoder $\mathcal{D}$. The full sampling procedure is summarized in Appendix~\ref{supp:algo}.

\begin{table*}[t]
    \centering
    \caption{\textbf{Comparison of bidirectional motion-text generation performance on HumanML3D.}
    Best/second-best unified results are \textbf{bold}/\underline{underlined}.}
    \label{tab:h3d_sota}\label{tab:cr_joint}

    \setlength{\tabcolsep}{2.35pt}
    \renewcommand{\arraystretch}{1.04}
    \small

    \resizebox{\textwidth}{!}{%
        \begin{tabular}{@{}l|cccccc|ccccccc@{}}
            \toprule
            \multirow{2}{*}{Method}
            & \multicolumn{6}{c|}{T2M}
            & \multicolumn{7}{c}{M2T} \\
            \cmidrule(lr){2-7}
            \cmidrule(l){8-14}

            & R@1$\uparrow$
            & R@2$\uparrow$
            & R@3$\uparrow$
            & FID$\downarrow$
            & Div$\rightarrow$
            & MM Dist$\downarrow$
            & R@1$\uparrow$
            & R@3$\uparrow$
            & BLEU@1$\uparrow$
            & BLEU@4$\uparrow$
            & ROUGE-L$\uparrow$
            & CIDEr$\uparrow$
            & BERTScore$\uparrow$ \\
            \midrule

            Real
            & 0.511\textsuperscript{$\pm 0.003$} & 0.703\textsuperscript{$\pm 0.003$} & 0.797\textsuperscript{$\pm 0.002$} & 0.002\textsuperscript{$\pm 0.000$} & 9.503\textsuperscript{$\pm 0.065$} & 2.974\textsuperscript{$\pm 0.008$}
            & 0.523 & 0.828 & -- & -- & -- & -- & -- \\

            \midrule
            \rowcolor{gray!10}
            \multicolumn{14}{c}{\textit{T2M-Only Models}} \\
            \midrule

            MDM~\cite{mdm}
            & -- & -- & 0.611\textsuperscript{$\pm 0.007$} & 0.544\textsuperscript{$\pm 0.044$} & 9.559\textsuperscript{$\pm 0.086$} & 5.566\textsuperscript{$\pm 0.027$}
            & -- & -- & -- & -- & -- & -- & --  \\

            MotionDiffuse~\cite{motiondiffuse}
            & 0.491\textsuperscript{$\pm 0.001$} & 0.681\textsuperscript{$\pm 0.001$} & 0.782\textsuperscript{$\pm 0.001$} & 0.630\textsuperscript{$\pm 0.001$} & 9.410\textsuperscript{$\pm 0.049$} & 3.113\textsuperscript{$\pm 0.001$}
            & -- & -- & -- & -- & -- & -- & --  \\

            MLD~\cite{mld}
            & 0.481\textsuperscript{$\pm 0.003$} & 0.673\textsuperscript{$\pm 0.003$} & 0.772\textsuperscript{$\pm 0.002$} & 0.473\textsuperscript{$\pm 0.013$} & 9.724\textsuperscript{$\pm 0.082$} & 3.196\textsuperscript{$\pm 0.010$}
            & -- & -- & -- & -- & -- & -- & --  \\

            MoMask~\cite{momask}
            & 0.521\textsuperscript{$\pm 0.002$} & 0.713\textsuperscript{$\pm 0.002$} & 0.807\textsuperscript{$\pm 0.002$} & 0.045\textsuperscript{$\pm 0.002$} & 9.620 & 2.958\textsuperscript{$\pm 0.008$}
            & -- & -- & -- & -- & -- & -- & --  \\

            T2M-GPT~\cite{t2mgpt}
            & 0.492\textsuperscript{$\pm 0.003$} & 0.679\textsuperscript{$\pm 0.002$} & 0.775\textsuperscript{$\pm 0.002$} & 0.141\textsuperscript{$\pm 0.005$} & 9.722\textsuperscript{$\pm 0.082$} & 3.121\textsuperscript{$\pm 0.009$}
            & -- & -- & -- & -- & -- & -- & --  \\

            ReMoDiffuse~\cite{remodiffuse}
            & 0.510\textsuperscript{$\pm 0.005$} & 0.698\textsuperscript{$\pm 0.006$} & 0.795\textsuperscript{$\pm 0.004$} & 0.103\textsuperscript{$\pm 0.004$} & 9.018\textsuperscript{$\pm 0.075$} & 2.974\textsuperscript{$\pm 0.016$}
            & -- & -- & -- & -- & -- & -- & --  \\

            MoGenTS~\cite{mogents}
            & 0.529\textsuperscript{$\pm 0.003$} & 0.719\textsuperscript{$\pm 0.002$} & 0.812\textsuperscript{$\pm 0.002$} & 0.033\textsuperscript{$\pm 0.001$} & 9.570\textsuperscript{$\pm 0.077$} & 2.867\textsuperscript{$\pm 0.006$} & -- & -- & -- & -- & -- & -- & --  \\

            MotionLCM~\cite{motionlcm}
            & 0.502\textsuperscript{$\pm 0.003$} & 0.698\textsuperscript{$\pm 0.002$} & 0.798\textsuperscript{$\pm 0.002$} & 0.304\textsuperscript{$\pm 0.012$} & 9.607\textsuperscript{$\pm 0.066$} & 3.012\textsuperscript{$\pm 0.007$}
            & -- & -- & -- & -- & -- & -- & --  \\

            BAMM~\cite{pinyoanuntapong2024bamm} & 0.525\textsuperscript{$\pm 0.002$} & -- & 0.814\textsuperscript{$\pm 0.003$} & 0.055\textsuperscript{$\pm 0.002$} & -- & 2.919 & -- & -- & -- & -- & -- & -- & -- \\

            BiPO~\cite{bipo} & 0.523\textsuperscript{$\pm 0.003$} & -- & 0.809\textsuperscript{$\pm 0.002$} & 0.030\textsuperscript{$\pm 0.002$} & -- & 2.880\textsuperscript{$\pm 0.009$} & -- & -- & -- & -- & -- & -- & -- \\

            CoMo~\cite{Huang2024CoMo}
            & 0.502\textsuperscript{$\pm 0.002$} & 0.692\textsuperscript{$\pm 0.007$} & 0.790\textsuperscript{$\pm 0.002$} & 0.262\textsuperscript{$\pm 0.004$} & 9.936\textsuperscript{$\pm 0.066$} & 3.032\textsuperscript{$\pm 0.015$}
            & -- & -- & -- & -- & -- & -- & --  \\

            MotionMamba~\cite{zhang2024motion}
            & 0.502\textsuperscript{$\pm 0.003$} & 0.693\textsuperscript{$\pm 0.002$} & 0.792\textsuperscript{$\pm 0.002$} & 0.281\textsuperscript{$\pm 0.009$} & 9.871\textsuperscript{$\pm 0.084$} & 3.060\textsuperscript{$\pm 0.058$}
            & -- & -- & -- & -- & -- & -- & --  \\

            EnergyMoGen~\cite{zhang2025energymogen}
            & 0.526\textsuperscript{$\pm 0.003$} & 0.718\textsuperscript{$\pm 0.003$} & 0.815\textsuperscript{$\pm 0.002$} & 0.176\textsuperscript{$\pm 0.006$} & 9.500\textsuperscript{$\pm 0.091$} & 2.931\textsuperscript{$\pm 0.007$}
            & -- & -- & -- & -- & -- & -- & --  \\
            \midrule
            \rowcolor{gray!10}
            \multicolumn{14}{c}{\textit{Separated Bidirectional Models}} \\
            \midrule

            TM2T~\cite{tm2t}
            & 0.424\textsuperscript{$\pm 0.003$} & 0.618\textsuperscript{$\pm 0.003$} & 0.729\textsuperscript{$\pm 0.002$} & 1.501\textsuperscript{$\pm 0.017$} & 8.589\textsuperscript{$\pm 0.076$} & 3.467\textsuperscript{$\pm 0.011$}
            & 0.516 & 0.823 & 48.9 & 7.0 & 38.1 & 16.8 & 32.2  \\

            LaMP~\cite{lamp}
            & 0.557\textsuperscript{$\pm 0.003$} & 0.751\textsuperscript{$\pm 0.002$} & 0.843\textsuperscript{$\pm 0.001$} & 0.032\textsuperscript{$\pm 0.002$} & 9.571\textsuperscript{$\pm 0.069$} & 2.759\textsuperscript{$\pm 0.007$} & 0.547 & 0.831 & 47.8 & 13.0 & 37.1 & 28.9 & --  \\

            MG-Mo.LLM~\cite{mgmotionllm}
            & 0.516\textsuperscript{$\pm 0.002$} & 0.706\textsuperscript{$\pm 0.002$} & 0.802\textsuperscript{$\pm 0.003$} & 0.303\textsuperscript{$\pm 0.010$} & 9.960\textsuperscript{$\pm 0.073$} & 2.952\textsuperscript{$\pm 0.009$} & 0.592 & 0.866 & -- & 8.1 & -- & -- & 36.7  \\

            \midrule
            \rowcolor{gray!10}
            \multicolumn{14}{c}{\textit{Unified Bidirectional Models}} \\
            \midrule

            MotionGPT~\cite{mgpt}
            & 0.492\textsuperscript{$\pm 0.003$} & 0.681\textsuperscript{$\pm 0.003$} & 0.778\textsuperscript{$\pm 0.002$} & 0.232\textsuperscript{$\pm 0.008$} & \underline{9.528}\textsuperscript{$\pm 0.071$} & 3.096\textsuperscript{$\pm 0.008$}
            & 0.543 & 0.827 & 48.2 & 12.5 & 37.4 & 29.2 & 32.4  \\

            MotionGPT2~\cite{mgpt2}
            & 0.496\textsuperscript{$\pm 0.002$} & 0.691\textsuperscript{$\pm 0.003$} & 0.782\textsuperscript{$\pm 0.004$} & 0.191\textsuperscript{$\pm 0.004$} & 9.860\textsuperscript{$\pm 0.026$} & 3.080\textsuperscript{$\pm 0.013$}
            & 0.558 & 0.838 & 48.7 & 13.8 & 37.6 & 29.8 & 32.6  \\

            MotionGPT3~\cite{mgpt3}
            & \underline{0.553}\textsuperscript{$\pm 0.003$} & \textbf{0.747}\textsuperscript{$\pm 0.002$} & \underline{0.837}\textsuperscript{$\pm 0.003$} & 0.208\textsuperscript{$\pm 0.006$} & 9.700\textsuperscript{$\pm 0.096$} & \textbf{2.725}\textsuperscript{$\pm 0.008$}
            & \underline{0.573} & 0.864 & 59.1 & 19.4 & \underline{46.2} & 28.7 & 35.2  \\

            MoTe~\cite{mote}
            & 0.548\textsuperscript{$\pm 0.002$} & 0.737\textsuperscript{$\pm 0.002$} & 0.825\textsuperscript{$\pm 0.002$} & 0.075\textsuperscript{$\pm 0.004$} & -- & 2.867\textsuperscript{$\pm 0.012$}
            & \textbf{0.577} & \textbf{0.871} & 46.7 & 11.2 & 37.4 & 31.5 & 30.3  \\

            DiMo~\cite{dimo}
            & 0.528\phantom{\textsuperscript{$\pm 0.000$}} & 0.724\phantom{\textsuperscript{$\pm 0.000$}} & 0.818\phantom{\textsuperscript{$\pm 0.000$}} & \textbf{0.047}\phantom{\textsuperscript{$\pm 0.000$}} & 9.419\phantom{\textsuperscript{$\pm 0.000$}} & 2.862\phantom{\textsuperscript{$\pm 0.000$}}
            & \textbf{0.577} & 0.855 & \textbf{64.2} & \textbf{22.7} & \textbf{47.1} & \underline{58.1} & \underline{37.7}  \\

            \rowcolor{gray!18}
            BiMoGen (Ours)
            & \textbf{0.555}\textsuperscript{$\pm 0.003$} & \underline{0.744}\textsuperscript{$\pm 0.003$} & \textbf{0.841}\textsuperscript{$\pm 0.002$} & \underline{0.069}\textsuperscript{$\pm 0.007$} & \textbf{9.524}\textsuperscript{$\pm 0.079$} & \underline{2.733}\textsuperscript{$\pm 0.008$} & \underline{0.573} & \underline{0.866} & \underline{60.1} & \underline{20.1} & 44.4 & \textbf{60.2} & \textbf{38.5}  \\

            \bottomrule
        \end{tabular}%
    }
    
\end{table*}
\begin{table*}[t]  
    \centering  
    \caption{\textbf{Comparison of bidirectional motion-text generation performance on KIT-ML.}  
    We report Text-to-Motion (T2M) and Motion-to-Text (M2T) results across two method categories.  
    The best results within unified models are highlighted in \textbf{bold}, and the second-best are \underline{underlined}.}  
    \label{tab:kit_sota}  
  
    \setlength{\tabcolsep}{2.35pt}  
    \renewcommand{\arraystretch}{1.04}  
    \small  
  
    \resizebox{\textwidth}{!}{%
        \begin{tabular}{@{}l|cccccc|ccccccc@{}}  
            \toprule  
            \multirow{2}{*}{Method}  
            & \multicolumn{6}{c|}{T2M}  
            & \multicolumn{7}{c}{M2T} \\  
            \cmidrule(lr){2-7}  
            \cmidrule(l){8-14}  
  
            & R@1$\uparrow$  
            & R@2$\uparrow$  
            & R@3$\uparrow$  
            & FID$\downarrow$  
            & Div$\rightarrow$  
            & MM Dist$\downarrow$  
            & R@1$\uparrow$  
            & R@3$\uparrow$  
            & BLEU@1$\uparrow$  
            & BLEU@4$\uparrow$  
            & ROUGE-L$\uparrow$  
            & CIDEr$\uparrow$  
            & BERTScore$\uparrow$ \\  
            \midrule

            Real  
            & 0.424\textsuperscript{$\pm 0.005$} & 0.649\textsuperscript{$\pm 0.006$} & 0.779\textsuperscript{$\pm 0.006$} & 0.031\textsuperscript{$\pm 0.004$} & 11.080\textsuperscript{$\pm 0.097$} & 2.788\textsuperscript{$\pm 0.012$}  
            & 0.399 & 0.793 & -- & -- & -- & -- & -- \\  
  
            \midrule  
            \rowcolor{gray!10}  
            \multicolumn{14}{c}{\textit{Separated Bidirectional Models}} \\  
            \midrule  
  
            TM2T~\cite{tm2t}  
            & 0.280\textsuperscript{$\pm 0.005$} & 0.463\textsuperscript{$\pm 0.006$} & 0.587\textsuperscript{$\pm 0.005$} & 3.599\textsuperscript{$\pm 0.153$} & 9.473\textsuperscript{$\pm 0.117$} & 4.591\textsuperscript{$\pm 0.026$} & 0.359 & 0.668 & 46.7 & 18.4 & 44.2 & 79.5 & 23.0  \\
  
            LaMP~\cite{lamp}  
            & 0.479\textsuperscript{$\pm 0.006$} & 0.691\textsuperscript{$\pm 0.005$} & 0.826\textsuperscript{$\pm 0.005$} & 0.141\textsuperscript{$\pm 0.013$} & 10.929\textsuperscript{$\pm 0.101$} & 2.704\textsuperscript{$\pm 0.018$} & 0.540 & 0.844 & -- & -- & -- & -- & --  \\

            \midrule  
            \rowcolor{gray!10}  
            \multicolumn{14}{c}{\textit{Unified Bidirectional Models}} \\  
            \midrule  
  
            MotionGPT~\cite{mgpt}  
            & 0.366\textsuperscript{$\pm 0.005$} & 0.558\textsuperscript{$\pm 0.004$} & 0.680\textsuperscript{$\pm 0.005$} & 0.510\textsuperscript{$\pm 0.016$} & 10.350\textsuperscript{$\pm 0.084$} & 3.527\textsuperscript{$\pm 0.021$}  
            & -- & -- & -- & -- & -- & -- & --  \\
  
            MotionGPT2~\cite{mgpt2}  
            & \textbf{0.427}\textsuperscript{$\pm 0.003$} & \underline{0.627}\textsuperscript{$\pm 0.002$} & \textbf{0.764}\textsuperscript{$\pm 0.003$} & 0.614\textsuperscript{$\pm 0.005$} & \textbf{11.256}\textsuperscript{$\pm 0.026$} & 3.164\textsuperscript{$\pm 0.013$}  
            & -- & -- & -- & -- & -- & -- & --  \\
  
            MoTe~\cite{mote}  
            & 0.419\textsuperscript{$\pm 0.005$} & \underline{0.627}\textsuperscript{$\pm 0.005$} & \underline{0.741}\textsuperscript{$\pm 0.004$} & 0.256\textsuperscript{$\pm 0.018$} & -- & 3.216\textsuperscript{$\pm 0.036$}  
            & \textbf{0.421} & \textbf{0.765} & 44.9 & 14.1 & 41.8 & 55.6 & 35.9  \\
  
            DiMo~\cite{dimo}  
            & 0.406\phantom{\textsuperscript{$\pm 0.000$}} & 0.620\phantom{\textsuperscript{$\pm 0.000$}} & \underline{0.741}\phantom{\textsuperscript{$\pm 0.000$}} & \textbf{0.206}\phantom{\textsuperscript{$\pm 0.000$}} & \underline{10.892}\phantom{\textsuperscript{$\pm 0.000$}} & \underline{2.983}\phantom{\textsuperscript{$\pm 0.000$}}  
            & 0.396 & 0.723 & \textbf{52.5} & \textbf{17.8} & \textbf{48.0} & \textbf{68.7} & \underline{37.7}  \\

            \rowcolor{gray!18}  
            BiMoGen (Ours) 
            & \underline{0.421}\textsuperscript{$\pm 0.004$} & \textbf{0.632}\textsuperscript{$\pm 0.007$} & 0.733\textsuperscript{$\pm 0.006$} & \underline{0.222}\textsuperscript{$\pm 0.019$} & 10.880\textsuperscript{$\pm 0.096$} & \textbf{2.763}\textsuperscript{$\pm 0.022$} & \underline{0.405} & \underline{0.734} & \underline{49.8} & \underline{17.3} & \underline{46.1} & \underline{62.2} & \textbf{40.4}  \\
  
            \bottomrule  
        \end{tabular}%
    }  
    
\end{table*}

\section{Experiments}\label{sec4}

\subsection{Experimental Setup}

\noindent \textbf{Datasets.} We evaluate BiMoGen mainly on HumanML3D~\cite{humanml3d}, a standard benchmark for bidirectional text and motion generation. HumanML3D contains 14{,}616 motion sequences from AMASS~\cite{amass} and HumanAct12~\cite{humanact12}, paired with 44{,}970 textual descriptions. We also report results on KIT-ML~\cite{plappert2016kit}, a smaller benchmark with 3{,}911 motion sequences and 6{,}278 textual annotations. For both datasets, we follow the settings and motion representations adopted in prior work~\cite{mgpt,mld}.

\noindent \textbf{Evaluation Protocol.}
Following the evaluation protocol of~\citet{humanml3d}, we use the pretrained motion-text feature extractor and average results over 20 evaluation runs. For text-to-motion (T2M) we report Top-$k$ retrieval accuracy (R@$k$), Fr\'echet Inception Distance (FID), Diversity, and MM Dist, the average feature-space distance between paired text and generated motion. For motion-to-text (M2T) we report retrieval accuracy R@1/R@3 against ground-truth motions, together with BLEU, ROUGE-L, CIDEr, and BERTScore for caption quality~\cite{cider,rouge,bertscore,bleu}. See Appendix~\ref{sec:app:metrics} for details.

\noindent \textbf{Implementation Details.} The bidirectional Transformer follows the LLaDA backbone~\cite{llada} with 20 layers and a hidden size of 1024. Motion tokens come from the single-layer VQ-VAE used in MotionGPT~\cite{mgpt,vqvae}, with a temporal downsampling rate of 4. We pretrain for 100 epochs at learning rate 2e-4 and fine-tune for 200 epochs at 8e-5, both with AdamW~\citep{adamw}. Fine-tuning batches mix T2M, M2T, and the auxiliary M2M objective at an 8:1:1 ratio, with 10\% condition dropout for CFG~\citep{cfg}. The self-correction weight $\lambda$ is set to 1. At inference we use 20 sampling steps, CFG scale for T2M set to 4, and self-correction at $\mathcal{R}{=}\{T/4, T/2\}$. We train and evaluate on a single NVIDIA H100.

\begin{table*}[t]
    \caption{Ablation on the training strategy on HumanML3D. PT denotes masked pretraining for cross-modal correspondence, SFT denotes supervised fine-tuning for bidirectional generation, and SC denotes the self-correction objective. \textbf{Bold} denotes the best performance.}
    \label{tab:ab_train_strategy}
    \centering
    \setlength{\tabcolsep}{3.0pt}
    \renewcommand{\arraystretch}{1.15}
    \small
    \resizebox{\textwidth}{!}{%
    \begin{tabular}{@{}c c c | c c c c c c | c c c c c c c@{}}
        \toprule
        \multicolumn{3}{c|}{Training Strategy}
        & \multicolumn{6}{c|}{T2M}
        & \multicolumn{7}{c}{M2T} \\
        \cmidrule(lr){1-3} \cmidrule(lr){4-9} \cmidrule(l){10-16}
        PT & SFT & SC
        & R@1 $\uparrow$ & R@2 $\uparrow$ & R@3 $\uparrow$ & FID $\downarrow$ & Div $\rightarrow$ & MM Dist $\downarrow$
        & R@1 $\uparrow$ & R@3 $\uparrow$ & BLEU@1 $\uparrow$ & BLEU@4 $\uparrow$ & ROUGE-L $\uparrow$ & CIDEr $\uparrow$ & BERTScore $\uparrow$ \\
        \midrule
        \multicolumn{3}{l|}{Real}
            & 0.511 & 0.703 & 0.797 & 0.002 & 9.503 & 2.974
            & 0.523 & 0.828 & -- & -- & -- & -- & -- \\

        \midrule    
        \cmark & \xmark & \xmark
        & 0.416 & 0.604 & 0.730 & 1.219 & 10.189 & 3.650
        & 0.434 & 0.736 & 24.3 & 5.5 & 26.4 & 6.7 & 16.8 \\
        \xmark & \cmark & \xmark
        & 0.466 & 0.633 & 0.726 & 0.325 & 9.546 & 3.259
        & 0.474 & 0.779 & 54.3 & 16.7 & 34.9 & 41.3 & 26.4 \\
        \cmark & \cmark & \xmark
        & 0.514 & 0.697 & 0.789 & 0.102 & 9.616 & 3.017
        & 0.547 & 0.832 & 56.1 & 16.4 & 42.9 & 44.0 & 36.7 \\
        \cmark & \cmark & \cmark
        & \textbf{0.555} & \textbf{0.744} & \textbf{0.841} & \textbf{0.069} & \textbf{9.524} & \textbf{2.733}
        & \textbf{0.573} & \textbf{0.866} & \textbf{60.1} & \textbf{20.1} & \textbf{44.4} & \textbf{60.2} & \textbf{38.5} \\
        \bottomrule
    \end{tabular}%
    }
\end{table*}

\begin{table*}[t]
    \centering
    \caption{Ablation of the trade-off between quality and computational cost. Representative bidirectional methods are included for reference under the same evaluation protocol. Lat. denotes Latency.}
    \label{tab:ab_inference}

    \setlength{\tabcolsep}{3.2pt}
    \renewcommand{\arraystretch}{1.08}
    \small

    \resizebox{\textwidth}{!}{%
        \begin{tabular}{@{}>{\centering\arraybackslash}m{2.8cm}|ccccc|cccccc|cc@{}}
            \toprule
            \multirow{2}{*}{Method}
            & \multicolumn{5}{c|}{Text-to-Motion}
            & \multicolumn{6}{c|}{Motion-to-Text}
            & \multicolumn{2}{c}{Computational Cost} \\
            \cmidrule(lr){2-6}
            \cmidrule(lr){7-12}
            \cmidrule(l){13-14}

            & R@1$\uparrow$
            & R@3$\uparrow$
            & FID$\downarrow$
            & MM Dist$\downarrow$
            & Lat. (s)$\downarrow$
            & R@1$\uparrow$
            & R@3$\uparrow$
            & BLEU@1$\uparrow$
            & BLEU@4$\uparrow$
            & ROUGE-L$\uparrow$
            & BERTScore$\uparrow$
            & \#Params
            & FLOPs \\

            \midrule
            MotionGPT~\cite{mgpt}
            & 0.492 & 0.778 & 0.232 & 3.096 & 1.04
            & 0.543 & 0.827 & 48.2 & 12.5 & 37.4 & 32.4
            & 220M & 7.45T \\

            MotionGPT3~\cite{mgpt3}
            & \underline{0.553} & 0.837 & 0.208 & \textbf{2.725} & 1.02
            & \underline{0.573} & 0.864 & 59.1 & 19.4 & \underline{46.2} & 35.2
            & 238M & 11.00T \\

            MG-Mo.LLM~\cite{mgmotionllm}
            & 0.516 & 0.802 & 0.303 & 2.952 & 1.09
            & \textbf{0.592} & \textbf{0.866} & -- & 8.1 & -- & 36.7
            & 220M & 1.66T \\

            DiMo (20 steps)~\cite{dimo}
            & 0.528 & 0.818 & \textbf{0.050} & 2.862 & 1.55
            & 0.568 & 0.845 & \textbf{62.5} & \textbf{22.0} & \textbf{47.3} & 35.4
            & 473M & 2.56T \\

            \midrule
            Ours (5 steps)
            & 0.533 & 0.832 & 0.118 & 2.787 & \textbf{0.20}
            & 0.467 & 0.749 & 57.6 & 16.8 & 44.3 & 28.9
            & 334M & 0.34T \\

            Ours (10 steps)
            & 0.543 & \underline{0.840} & 0.082 & 2.774 & \underline{0.35}
            & 0.525 & 0.823 & 59.1 & 17.3 & 44.8 & 33.3
            & 334M & 0.59T \\

            Ours (20 steps)
            & \textbf{0.555} & \textbf{0.841} & \underline{0.069} & \underline{2.733} & 0.67
            & \underline{0.573} & \textbf{0.866} & \underline{60.1} & \underline{20.1} & 44.4 & \underline{38.5}
            & 334M & 1.08T \\

            Ours (30 steps)
            & {0.550} & \underline{0.840} & 0.071 & 2.756 & 0.95
            & \underline{0.573} & 0.853 & 59.2 & 19.0 & 46.0 & \textbf{41.2}
            & 334M & 1.57T \\

            \bottomrule
        \end{tabular}%
    }
    
\end{table*}

\begin{figure}[t]
    \centering
    \includegraphics[width=\linewidth]{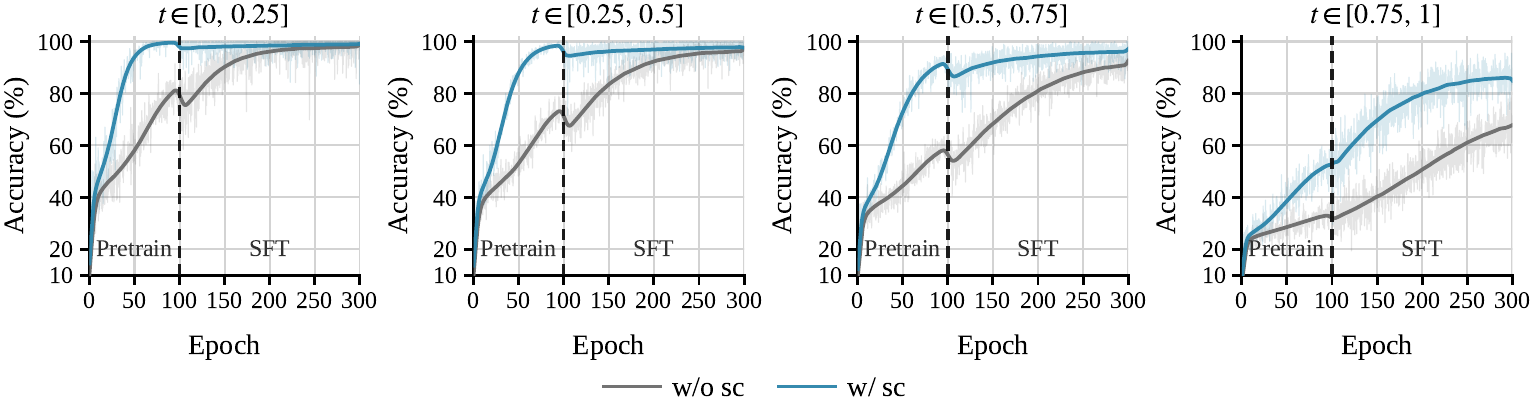}
    \caption{Training accuracy under different masking levels $t$. Each panel shows one masking interval. The dashed vertical line separates masked pretraining and supervised fine-tuning.}
    \label{fig:mask_ratio_acc}
    
\end{figure}
\begin{figure}[t]
    \centering
    \includegraphics[width=\linewidth]{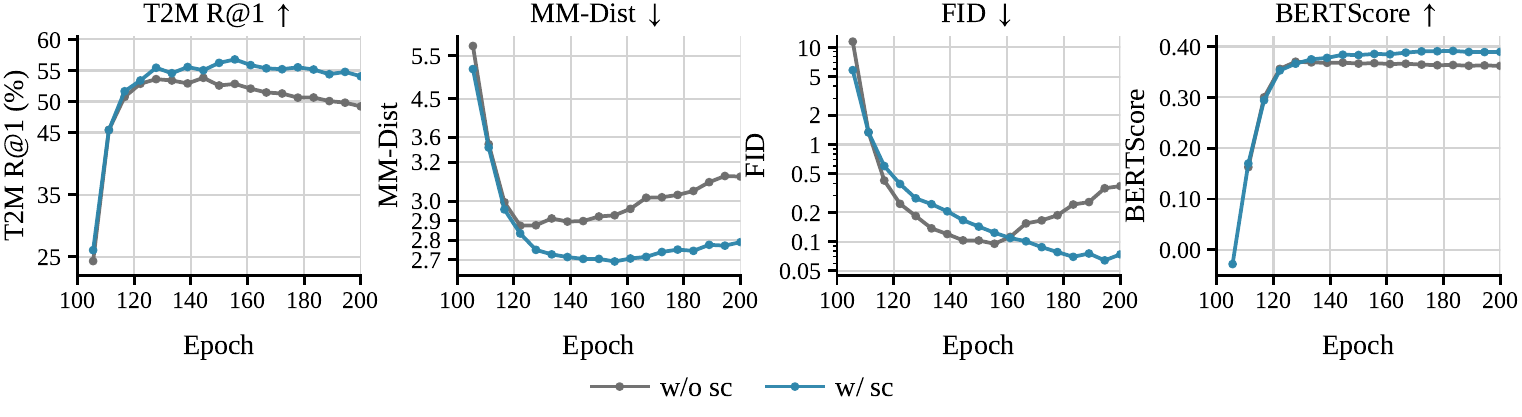}
    \caption{SFT evaluation over epochs. The first three panels correspond to T2M, and the last panel corresponds to M2T. Curves compare training with and without self-correction.}
    \label{fig:performance}
\end{figure}

\begin{table}[t]
\centering
\caption{Training and sampling self-correction on HumanML3D. All variants use PT and SFT. R@1, FID, and MM Dist evaluate T2M; CIDEr evaluates M2T. Sampling SC uses $\mathcal{R}=\{T/4,T/2\}$.}
\label{tab:cr_sc_factorial}
\small
\setlength{\tabcolsep}{6pt}
\renewcommand{\arraystretch}{1.12}
\begin{adjustbox}{max width=\linewidth,center}
\begin{tabular}{@{}ccrrrr@{}}
\toprule
Training SC & Sampling SC & R@1 $\uparrow$ & FID $\downarrow$ & MM Dist $\downarrow$ & CIDEr $\uparrow$ \\
\midrule
\xmark & \xmark & 0.514 & 0.102 & 3.017 & 44.0 \\
\xmark & \cmark & 0.503 & 0.118 & 3.132 & 42.9 \\
\cmark & \xmark & 0.551 & 0.070 & 2.745 & 58.8 \\
\cmark & \cmark & 0.555 & 0.069 & 2.733 & 60.2 \\
\bottomrule
\end{tabular}
\end{adjustbox}
\end{table}

\begin{table*}[t]
    \centering
    \caption{Ablation on the SC schedule during sampling on HumanML3D. \textit{Disabled}  removes SC during sampling only. All variants use the same number of sampling steps $T$. For T2M, the CFG scale $w$ is fixed to its default value. CFG is not used for M2T. Lat. denotes Latency.}
    \label{tab:sc_schedule}
    \setlength{\tabcolsep}{2.2pt}
    \renewcommand{\arraystretch}{1.06}
    \small

    \resizebox{\textwidth}{!}{%
        \begin{tabular}{@{}l|cccccc|ccccccc@{}}
            \toprule
            \multirow{2}{*}{SC Schedule $\mathcal{R}$}
            & \multicolumn{6}{c|}{T2M}
            & \multicolumn{7}{c}{M2T} \\
            \cmidrule(lr){2-7}
            \cmidrule(l){8-14}

            & R@1$\uparrow$
            & R@3$\uparrow$
            & FID$\downarrow$
            & Div$\rightarrow$
            & MM Dist$\downarrow$
            & Lat. (s)$\downarrow$
            & R@1$\uparrow$
            & R@3$\uparrow$
            & BLEU@1$\uparrow$
            & BLEU@4$\uparrow$
            & ROUGE-L$\uparrow$
            & CIDEr$\uparrow$
            & BERTScore$\uparrow$ \\

            \midrule
            \textit{Disabled}
            & 0.551 & \underline{0.838} & 0.070 & 9.870 & \underline{2.745} & \textbf{0.55}
            & 0.568 & 0.864 & 59.2 & 18.9 & \underline{45.6} & 58.8 & \underline{39.6} \\

            \midrule
            $T/4$
            & \underline{0.553} & 0.836 & 0.067 & 9.397 & 2.764 & \underline{0.61}
            & 0.570 & \underline{0.866} & 59.2 & 19.0 & \underline{45.6} & 58.8 & 38.6 \\

            $T/2$
            & 0.551 & 0.837 & 0.070 & 9.697 & 2.782 & \underline{0.61}
            & 0.570 & \underline{0.866} & 59.2 & 18.9 & \underline{45.6} & 58.9 & 38.6 \\

            $3T/4$
            & 0.542 & 0.831 & \textbf{0.056} & \underline{9.423} & 2.833 & \underline{0.61}
            & 0.570 & \underline{0.866} & 59.2 & 18.9 & \underline{45.6} & 59.6 & 38.8 \\

            Every step
            & 0.529 & 0.814 & \underline{0.066} & 9.570 & 2.948 & 0.78
            & \textbf{0.580} & \textbf{0.868} & \textbf{60.2} & \underline{19.7} & \textbf{45.8} & \textbf{61.2} & \textbf{41.0} \\
            
            \midrule
            $T/4,\,T/2$
            & \textbf{0.555} & \textbf{0.841} & 0.069 & \textbf{9.524} & \textbf{2.733} & 0.67
            & \underline{0.573} & \underline{0.866} & \underline{60.1} & \textbf{20.1} & 44.4 & \underline{60.2} & 38.5 \\
            \bottomrule
        \end{tabular}%
    }
\end{table*}

\subsection{Main Results}

\noindent \textbf{Comparisons on Bidirectional Motion-Text Generation.} Table~\ref{tab:h3d_sota} reports the main results on HumanML3D. Because BiMoGen is designed as a unified model for both T2M and M2T, we focus our comparisons on the unified bidirectional setting, while using T2M-Only and separated bidirectional methods only as references. On T2M, BiMoGen achieves the best retrieval performance among unified models, with an R@1 of 0.555 and an R@3 of 0.841. It also maintains competitive motion quality, with an FID of 0.069 and an MM Dist of 2.733. These results show that masked bidirectional prediction improves motion-text alignment over autoregressive unified models such as MotionGPT and MotionGPT3~\cite{mgpt,mgpt3}, while still producing realistic motions. DiMo~\cite{dimo} achieves lower FID, but differences in tokenization, architecture, and training preclude attributing this gap solely to tokenization.  On M2T, BiMoGen achieves the highest CIDEr and BERTScore among unified models, reaching 60.2 and 38.5, respectively. This indicates a stronger ability in motion understanding, especially compared with autoregressive unified methods. Appendix~\ref{supp:single_task} compares unified and T2M-only training.

\begin{table}[t]
\centering
    \caption{Self-correction behavior across sampling steps. \textit{Modification} is the ratio of committed tokens revised by SC. \textit{Pred. Stability} is the ratio of revised tokens that remain unchanged at the next step. Toward-GT and Away-from-GT report the direction of changes among revised tokens.}
    \label{tab:sc_behavior}\label{tab:cr_gt}
    
    \setlength{\tabcolsep}{5pt}
    \renewcommand{\arraystretch}{1.15}
    \small
    \begin{adjustbox}{max width=0.95\linewidth,center}
    \begin{tabular}{@{}crrrr@{}}
        \toprule
        SC Step & Modification (\%) & Pred. Stability (\%) & Toward-GT (\%) & Away-from-GT (\%) \\
        \midrule
        $T/4$    & 31.82 & 84.70 & 68.5 & 31.5 \\
        $T/2$    & 29.48 & 83.72 & 64.7 & 35.3 \\
        $3T/4$   & 25.72 & 83.74 & 60.2 & 39.8 \\
        $T$      & 22.38 & 82.73 & 55.8 & 44.2 \\
        \bottomrule
    \end{tabular}%
    \end{adjustbox}
\end{table}
\begin{table*}[t]

    \centering

    \caption{T2M results with different motion tokenizers on HumanML3D. Part VQ-VAE follows ParCo~\cite{Zou2025ParCo}, which discretizes motion by body parts. VQ-VAE denotes the whole-body tokenizer~\cite{mgpt}. $^*$ denotes base and residual RVQ codes flattened into one sequence.}
    \label{tab:ab_tokenizer}
    \label{tab:cr_rvq}

    \setlength{\tabcolsep}{4.5pt}
    \renewcommand{\arraystretch}{1.08}
    \small

    \begin{adjustbox}{max width=\linewidth,center}
    \begin{tabular}{@{}lcccccc@{}}
        \toprule
        \multirow{2}{*}{Motion Tokenizer}
        & \multicolumn{6}{c}{Text-to-Motion Generation} \\
        \cmidrule(l){2-7}

        & R@1$\uparrow$
        & R@2$\uparrow$
        & R@3$\uparrow$
        & FID$\downarrow$
        & Div$\rightarrow$
        & MM Dist$\downarrow$ \\
        \midrule

        Real
        & 0.511 & 0.703 & 0.797 & 0.002 & 9.503 & 2.974 \\

        \midrule
        Part VQ-VAE~\cite{Zou2025ParCo}
        & 0.512 & 0.703 & 0.801 & 0.124 & 9.886 & 3.084 \\

        RVQ-VAE~\cite{momask}$^{*}$
        & 0.484 & 0.665 & 0.766 & 0.487 & 9.992 & 3.114 \\

        VQ-VAE~\cite{mgpt}
        & \textbf{0.555} & \textbf{0.744} & \textbf{0.841} & \textbf{0.069} & \textbf{9.524} & \textbf{2.733} \\

        \bottomrule
    \end{tabular}
    \end{adjustbox}\par
\end{table*}

\subsection{Ablation and Analysis}

\noindent\textbf{Effect of Training Strategy.} Table~\ref{tab:ab_train_strategy} isolates the contribution of each training stage. PT alone learns useful cross-modal correspondence, but its high FID and weak captioning scores show that correspondence pretraining is not sufficient for bidirectional generation. SFT is therefore necessary for producing valid target sequences. Adding PT before SFT improves alignment and motion quality. MM Dist decreases from 3.259 to 3.017, FID decreases from 0.325 to 0.102, and M2T R@1 increases from 0.474 to 0.547. These results support correspondence pretraining as an initialization for conditional generation.

{\textbf{Training and Sampling Self-Correction.} Table~\ref{tab:cr_sc_factorial} separates the two components. With sampling SC disabled, training SC raises R@1 from 0.514 to 0.551, lowers FID from 0.102 to 0.070, and improves CIDEr from 44.0 to 58.8. Sampling SC alone degrades these metrics. After training with SC, sampling correction further raises R@1 to 0.555 and CIDEr to 60.2. The main gain therefore comes from training on self-generated context, which enables further refinement at sampling time.} Figure~\ref{fig:mask_ratio_acc} shows the same effect in the training dynamics. At low mask ratios, SC mainly accelerates convergence and reaches a similar final accuracy. As the mask ratio increases, the advantage becomes larger and persists throughout training, especially at $t \in [0.75, 1]$. This regime matches the early stage of iterative sampling, where the model predicts from sparse context and then conditions on its previous outputs. Training with SC therefore reduces the mismatch between training on corrupted ground-truth context and inference on self-generated context, leading to more stable generation in both directions. Figure~\ref{fig:performance} further reports the SFT evaluation over epochs on both T2M and M2T, comparing training with and without SC.

\begin{figure}[!t]
    \centering
    \includegraphics[width=0.72\linewidth]{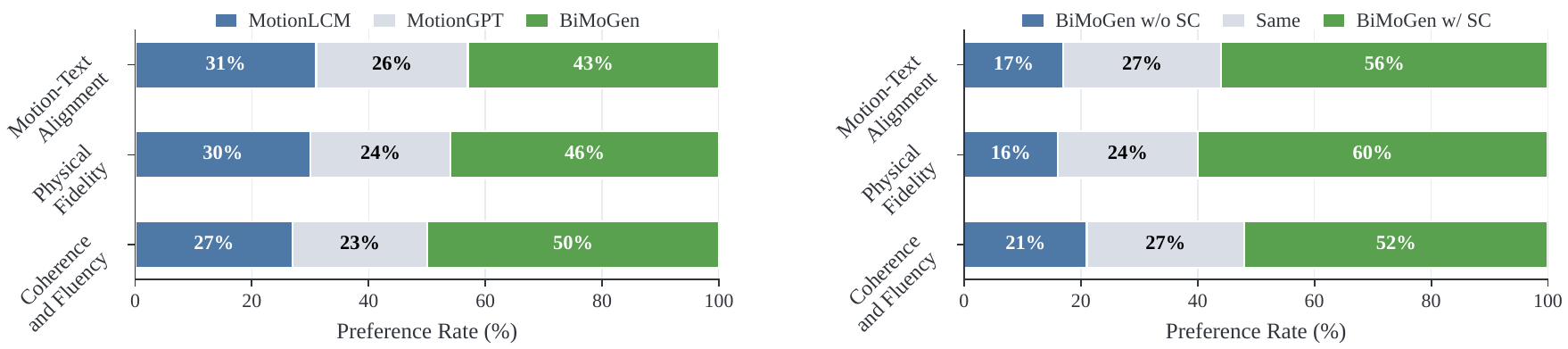}
    \caption{\textbf{User study.}
    Preference rates on 30 HumanML3D test prompts. Left: comparison with MotionLCM and MotionGPT. Right: ablation of SC with an additional ``Same'' option.}
\label{fig:user}

\end{figure}

\noindent {\textbf{Sampling Correction Schedule.}} Table~\ref{tab:sc_schedule} studies the timing of SC during masked sampling. Compared with \textit{Disabled}, applying SC at both $T/4$ and $T/2$ improves T2M alignment, increasing R@1 and R@3 to 0.555 and 0.841 and reducing MM Dist from 2.745 to 2.733, while maintaining a competitive FID of 0.069. It also improves M2T R@1, BLEU@4, and CIDEr over \textit{Disabled}, with a moderate latency increase from 0.55s to 0.67s. Single-step correction at $T/4$ or $T/2$ alone brings less consistent gains, while late correction at $3T/4$ achieves the lowest FID but weakens text-motion alignment, suggesting that late-stage revision mainly adjusts motion statistics rather than improving cross-modal consistency. Applying SC at every step further improves a few captioning metrics but degrades T2M retrieval. We therefore use $\mathcal{R}=\{T/4,T/2\}$ as the default.

Table~\ref{tab:sc_behavior} further analyzes the correction behavior. The modification rate decreases from 31.82\% at $T/4$ to 22.38\% at $T$, indicating more frequent revisions at earlier sampling steps. The stability of revised tokens remains high between 82.73\% and 84.70\%, so most revisions agree with the next step prediction. Corrections are more frequent at earlier sampling steps, and a larger fraction of the revised tokens move toward ground truth, suggesting that early correction is more effective at reducing prediction errors.

\noindent\textbf{Computation-Quality Trade-Off.} 
Table~\ref{tab:ab_inference} analyzes the effect of sampling steps and compares the computational cost with representative baselines. Increasing the number of steps from 5 to 20 improves retrieval in both generation directions. For T2M, FID decreases from 0.118 to 0.069. For M2T, retrieval accuracy also increases substantially, with BERTScore increasing from 28.9 to 38.5.  The setting with 20 steps gives the best overall balance. It attains the strongest T2M retrieval and the best R@3 among the compared bidirectional methods, while using 1.08T FLOPs. This is substantially lower than MotionGPT and DiMo, which require 7.45T FLOPs and 2.56T FLOPs. Since 30 steps brings no consistent gain, we use 20 denoising steps in the main experiments.

\noindent\textbf{Effect of Motion Tokenizer.}
We study the effect of motion tokenizer by replacing the whole-body VQ-VAE with the Part VQ-VAE from ParCo~\cite{Zou2025ParCo}, which quantizes different body parts with independent codebooks. As shown in Table~\ref{tab:ab_tokenizer}, this leads to worse performance. We attribute the drop to the increased learning difficulty introduced by part-wise tokenization under masked bidirectional generation. Part VQ-VAE produces multiple fine-grained tokens, and modeling their dependencies with full attention requires the model to learn both temporal dynamics and cross-part coordination simultaneously, which may be challenging under our current model capacity. {The flattened RVQ variant similarly introduces a longer sequence with dependencies across residual levels and yields an FID of 0.487 under the same single-Transformer design. Together, these comparisons support whole-body tokenization for the evaluated masked generation framework.}

\noindent\textbf{User Study.} As shown in Figure~\ref{fig:user}, BiMoGen achieves the highest preference rates across all three criteria. Adding SC further improves user preference, indicating better text alignment, physical plausibility, and temporal consistency. See Appendix~\ref{supp:user_study} for the study protocol.

\section{Conclusion}

We presented BiMoGen, a unified masked discrete diffusion framework for bidirectional motion-text modeling. By formulating both T2M and M2T as iterative masked prediction in a shared motion-text token space, BiMoGen avoids the rigid left-to-right order of autoregressive decoding and enables bidirectional conditioning between language and motion. DUCMT establishes cross-modal correspondence before conditional fine-tuning, while GASC trains recovery from self-generated context and enables further sampling refinement. Experiments on HumanML3D and KIT-ML show consistent gains in both directions, with competitive performance among unified models.

\noindent \textbf{Limitations.}\label{limi}
Our method employs a VQ-VAE motion tokenizer to convert continuous human motion into discrete tokens. As a result, generation quality is still partially bounded by the tokenizer's reconstruction fidelity and codebook expressiveness, especially for subtle or highly detailed motions. Moreover, the current formulation models motion at the sequence level and lacks explicit fine-grained control over individual body parts, such as hands, arms, or legs. Incorporating stronger part-aware tokenizers and more controllable representations is a promising direction for future work.

\section*{Acknowledgements}
This work was supported by National Science Foundation of China (62672109, 62302093, 52441503), Jiangsu Province Natural Science Fund (BK20230833) and the Open Research Fund of the State Key Laboratory of Multimodal Artificial Intelligence Systems under Grant E5SP060116. We thank the Big Data Computing Center of Southeast University for providing the facility support on the numerical calculations. 

\newpage
\bibliography{references}

\appendix


\clearpage
\appendix
\setcounter{section}{0}
\setcounter{figure}{0}
\setcounter{table}{0}
\setcounter{equation}{0}
\renewcommand\thesection{\Alph{section}}
\renewcommand{\theequation}{S\arabic{equation}}
\renewcommand{\thefigure}{S\arabic{figure}}
\renewcommand{\thetable}{S\arabic{table}}

\begin{center}
{\Large \textbf{BiMoGen: Bidirectional Motion-Text Generation via  \\[0.5em] Unified Masked Discrete Diffusion}} \\[0.5em]
{\large Supplementary Material}
\end{center}

\vspace{1em}
This supplementary document contains additional visualizations, experimental details, ablation results, and sampling pseudo-code for BiMoGen. It is structured as follows. Sec.~\ref{supp:vis} presents additional qualitative results on text-to-motion generation and motion in-between. Sec.~\ref{supp:exp} provides further experimental details (Sec.~\ref{supp:detail}) and ablation studies (Sec.~\ref{supp:ablation}), including the effects of semi-autoregressive sampling, backbones and classifier-free guidance, followed by the user study protocol (Sec.~\ref{supp:user_study}). Sec.~\ref{supp:algo} presents the full pseudo-code of sampling with self-correction. We discuss the broader impact of the proposed method in Sec.~\ref{supp:impact}.

\section{More Visualization}
\label{supp:vis}

In this section, we provide additional qualitative results to further demonstrate the effectiveness of BiMoGen. We compare BiMoGen with representative baselines on text-to-motion generation and motion in-between.

\begin{figure}[ht]
    \centering
    \includegraphics[width=\linewidth]{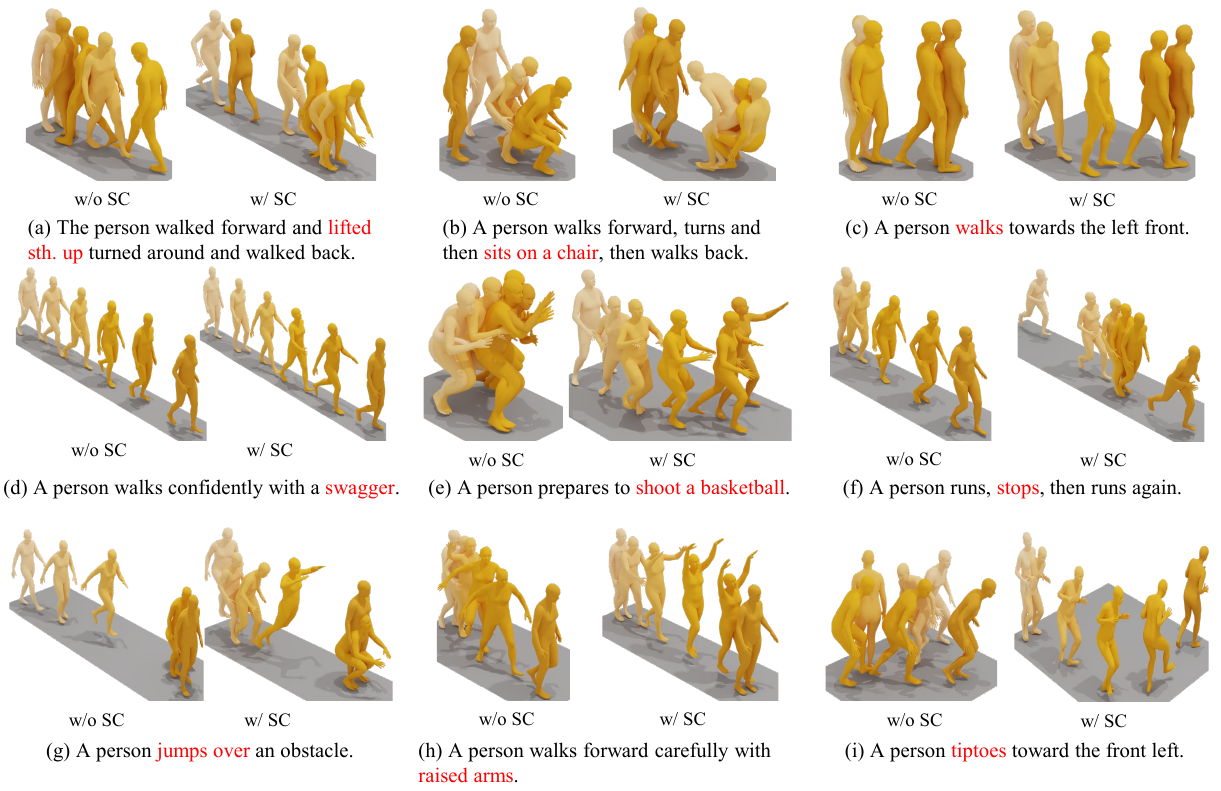}
    \caption{Qualitative comparison on text-to-motion generation. Red text marks descriptions that are not correctly reflected in the generated motion. BiMoGen produces motions that better align with the input prompts.}
    \label{supp:fig:t2m}
\end{figure}
\noindent\textbf{Visualization of Text-to-Motion Generation.}
Fig.~\ref{supp:fig:t2m} shows qualitative comparisons of text-to-motion generation on the HumanML3D test set. Compared with BiMoGen w/o SC, BiMoGen w/ SC produces motions that are more faithful to the input descriptions, especially in cases involving fine-grained semantics and directional cues.

\begin{figure}[ht]
    \centering
    \includegraphics[width=\linewidth]{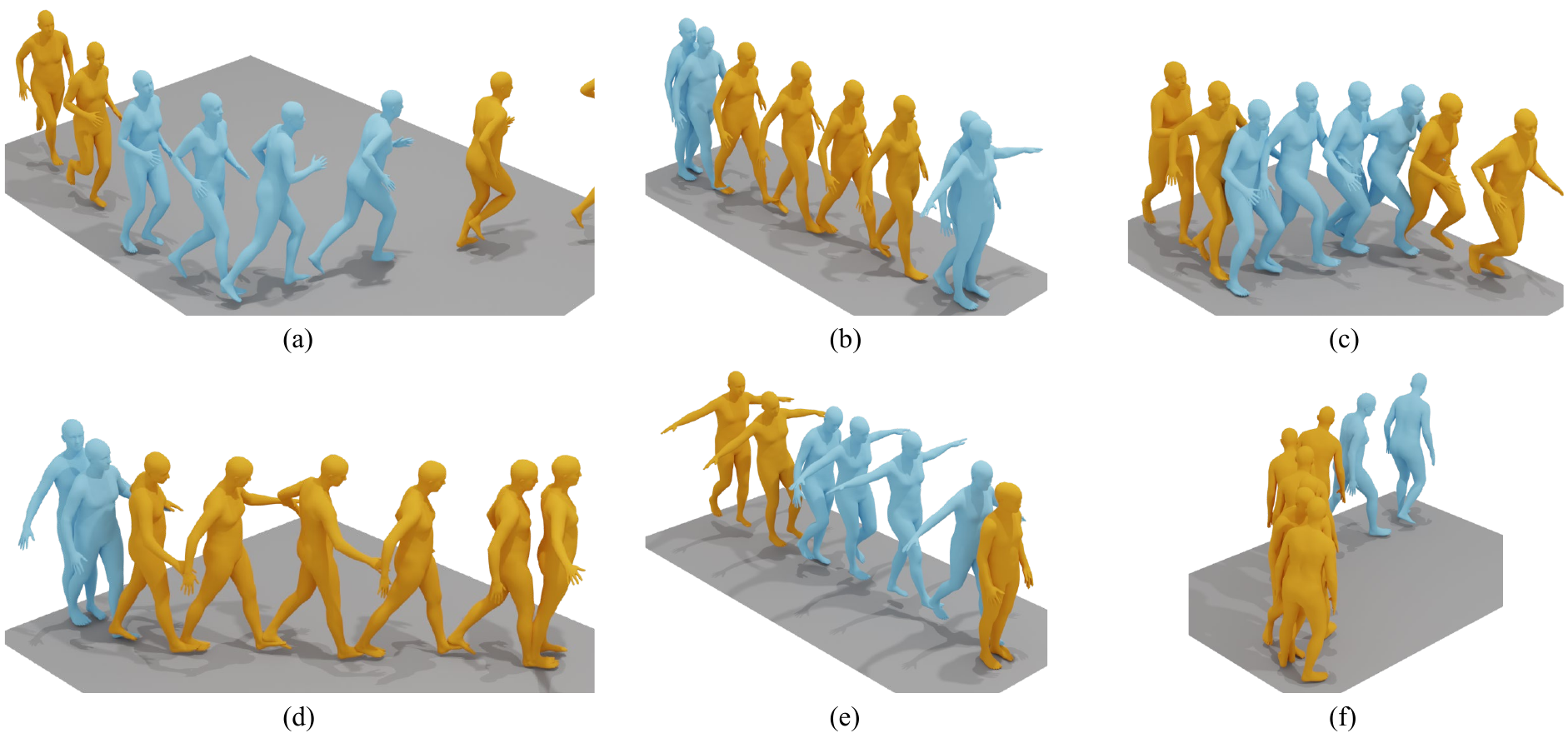}
    \caption{Visualization of motion in-betweening. Given prefix, suffix, middle, or start–end motion, BiMoGen performs motion in-betweening with smooth transitions. The human in blue represents the input, while the one in orange is generated by BiMoGen.}
    \label{supp:fig:inbetween}
\end{figure}

\noindent\textbf{Visualization of Motion In-between.}
Fig.~\ref{supp:fig:inbetween} shows motion in-between results, where the start and end segments are fixed and the middle portion is generated. The masked-diffusion formulation supports this setting without architectural change, and BiMoGen produces smooth and semantically consistent transitions.
\section{Experiments}
\label{supp:exp}

\subsection{More Experiment Details}
\label{supp:detail}

\noindent\textbf{Metrics Definition.}\label{sec:app:metrics}
We follow the evaluation protocol used in prior text and motion generation work~\citep{humanml3d, momask, tm2t, mgpt, mgpt3}. For T2M, we evaluate text and motion alignment, motion realism, and motion diversity. Unless otherwise noted, feature based metrics are computed with the official HumanML3D evaluator~\citep{humanml3d}, using motion encoder $\phi(\cdot)$ and text encoder $\psi(\cdot)$. For M2T, we follow prior captioning evaluation~\citep{tm2t} and adopt standard NLP metrics including BLEU~\citep{bleu}, ROUGE-L~\citep{rouge}, CIDEr~\citep{cider}, and BERTScore~\citep{bertscore} to evaluate the fluency, relevance, and diversity of generated captions. We also report retrieval based R Precision for M2T to measure the alignment between generated texts and the corresponding motions.

\noindent \textbf{Text and motion alignment.}
R@k measures retrieval accuracy in the shared evaluator space. For each query from one modality, candidates from the other modality are ranked by their evaluator distance, and the score is the fraction of samples whose paired item appears in the top $k$ results. MM Dist measures the average distance between paired text and motion embeddings.

\noindent \textbf{Motion realism.}
Fr\'echet Inception Distance measures the distributional distance between generated motions and ground-truth motions in the feature space. 

\noindent \textbf{Motion diversity.}
Diversity measures the variation among generated motions in the evaluator feature space. 

\noindent \textbf{Motion captioning.}
For M2T, generated descriptions are evaluated against reference descriptions. BLEU and ROUGE-L measure lexical overlap, CIDEr measures consensus with reference captions, and BERTScore measures embedding based semantic similarity. These metrics are used together because a single overlap metric does not fully capture caption quality.

\subsection{Additional Ablation Study}
\label{supp:ablation}

We provide additional ablation studies on the design choices of BiMoGen.

\subsubsection{Training Mixture}\label{supp:single_task}\label{supp:cr_m2m}
Table~\ref{tab:cr_m2m} compares T2M:M2T:M2M training ratios of 10:0:0, 8:1:1, and 8:2:0. T2M-only training (10:0:0) yields stronger T2M results than unified training (8:1:1), indicating a trade-off in supporting both generation directions. Removing M2M with an 8:2:0 mixture decreases R@1 from 0.555 to 0.522 and CIDEr from 60.2 to 58.8, while increasing FID from 0.069 to 0.133. Because the M2T proportion also changes, this comparison does not isolate M2M independently of the training mixture.
\begin{table}[t]
\centering
\caption{Effect of training mixtures on HumanML3D. Ratios denote T2M:M2T:M2M; 10:0:0 is T2M-only.}
\label{tab:cr_m2m}
\small
\setlength{\tabcolsep}{6pt}
\renewcommand{\arraystretch}{1.12}
\begin{adjustbox}{max width=\linewidth,center}
\begin{tabular}{@{}lrrrrrr@{}}
\toprule
T2M:M2T:M2M & R@1 $\uparrow$ & R@3 $\uparrow$ & FID $\downarrow$ & MM Dist $\downarrow$ & CIDEr $\uparrow$ & BERTScore $\uparrow$ \\
\midrule
10:0:0 & 0.562 & 0.852 & 0.047 & 2.649 & -- & -- \\
8:1:1 & 0.555 & 0.841 & 0.069 & 2.733 & 60.2 & 38.5 \\
8:2:0 & 0.522 & 0.807 & 0.133 & 3.004 & 58.8 & 37.2 \\
\bottomrule
\end{tabular}
\end{adjustbox}
\end{table}

\subsubsection{Backbone and Sampling Strategy}
\begin{table*}[t]
    \centering
    \caption{T2M results with different bidirectional Transformer backbones on HumanML3D. Bert-Base and Bert-Large use pretrained weights, while BiMoGen is trained from scratch with the same training strategy.}
    \label{tab:ab_backbone}
    
    \setlength{\tabcolsep}{2.35pt}
    \renewcommand{\arraystretch}{1.04}
    \small

    \resizebox{0.65\textwidth}{!}{%
        \begin{tabular}{@{}l|ccccccc@{}}
            \toprule
            \multirow{2}{*}{Backbone}
            & \multicolumn{7}{c}{T2M} \\
            \cmidrule(l){2-8}

            & \#Param
            & R@1$\uparrow$
            & R@2$\uparrow$
            & R@3$\uparrow$
            & FID$\downarrow$
            & Div$\rightarrow$
            & MM Dist$\downarrow$ \\
            \midrule

            Real
            & -- & 0.511 & 0.703 & 0.797 & 0.002 & 9.503 & 2.974 \\

            \midrule
            Bert-Base~\cite{bert}
            & 133M & 0.544 & 0.737 & 0.829 & 0.063 & 10.177 & 2.792 \\
            Bert-Large~\cite{bert}
            & 366M & \underline{0.553} & \textbf{0.747} & \underline{0.839} & \textbf{0.057} & \underline{10.018} & \underline{2.734} \\
            BiMoGen (Ours)
            & 334M & \textbf{0.555} & \underline{0.744} & \textbf{0.841} & \underline{0.069} & \textbf{9.524} & \underline{2.733} \\

            \bottomrule
        \end{tabular}%
    }
    \vspace{-5pt}
\end{table*}

\begin{table}[t]
    \centering
    \caption{Ablation on the CFG scale $w$ for T2M generation. The number of sampling steps $T$ is set to 20, and the self-correction is disabled.}
    \label{tab:ablation_cfg}

    \setlength{\tabcolsep}{3.2pt}
    \renewcommand{\arraystretch}{1.08}
    \small

    \resizebox{0.6\linewidth}{!}{%
        \begin{tabular}{@{}c|cccccc@{}}
            \toprule
            \multirow{2}{*}{CFG Scale $w$}
            & \multicolumn{6}{c}{Text-to-Motion} \\
            \cmidrule(l){2-7}

            & R@1$\uparrow$
            & R@2$\uparrow$
            & R@3$\uparrow$
            & FID$\downarrow$
            & Div$\rightarrow$
            & MM Dist$\downarrow$ \\

            \midrule

            1.0
            & 0.505 & 0.693 & 0.795 & 0.323 & 9.977 & 3.305 \\

            2.0
            & 0.535 & 0.729 & 0.827 & 0.144 & 10.001 & 2.840 \\

            3.0
            & \underline{0.550} & 0.\textbf{740} & 0.829 & 0.088 & 9.884 & \underline{2.753} \\

            4.0
            & \textbf{0.551} & \underline{0.739} & \textbf{0.838} & \textbf{0.070} & \underline{9.870} & \textbf{2.745} \\

            5.0
            & 0.543 & \underline{0.739} & \underline{0.833} & \underline{0.076} & \textbf{9.566} & 2.760 \\
            
            \bottomrule
        \end{tabular}%
    }
    \vspace{-10pt}
\end{table}

\begin{table}[t]
\centering
\caption{Effect of semi-autoregressive sampling for T2M generation. The first row denotes the standard parallel masked decoding used by BiMoGen. The remaining rows denote block-wise left-to-right semi-autoregressive variants, where tokens within each block are generated in parallel and different blocks are generated sequentially.}
\label{tab:ab_block}
\resizebox{0.8\linewidth}{!}{
\begin{tabular}{c|c|cccccc}
\toprule
Decoding Strategy & Block Size & R@1 $\uparrow$ & R@2 $\uparrow$ & R@3 $\uparrow$ & FID $\downarrow$ & Div $\rightarrow$ & MM Dist $\downarrow$ \\
\midrule
Parallel sampling & -- & \textbf{0.555} & \textbf{0.744} & \textbf{0.841} & \textbf{0.069} & \textbf{9.524} & \textbf{2.733} \\
\midrule
\multirow{5}{*}{Semi-Autoregressive sampling}
& 1  & 0.534 & 0.721 & 0.827 & 0.181 & 9.415 & 2.915 \\
& 2  & 0.521 & 0.716 & 0.821 & 0.220 & 9.640 & 2.939 \\
& 4  & 0.521 & 0.717 & 0.818 & 0.336 & 9.534 & 2.985 \\
& 5  & 0.511 & 0.693 & 0.787 & 0.527 & 9.435 & 3.040 \\
& 10 & 0.496 & 0.681 & 0.776 & 1.018 & 9.332 & 3.148 \\
\bottomrule
\end{tabular}
}
\vspace{-10pt}
\end{table}
\paragraph{Robustness across backbones.} Table~\ref{tab:ab_backbone} compares different bidirectional Transformer backbones under the same training recipe. The BERT variants are initialized from pretrained checkpoints, whereas BiMoGen is trained from scratch. Despite this difference, BiMoGen remains on par with BERT-Large and outperforms BERT-Base on most T2M metrics, with the best R@1, R@3, Div, and MM Dist. This result suggests that the proposed training strategy does not rely on language pretrained weights and can make effective use of model capacity across different backbone designs.

\noindent\textbf{Effect of classifier-free guidance.} Table~\ref{tab:ablation_cfg} studies the effect of the CFG scale $w$ for T2M generation. Increasing $w$ from 1.0 to 4.0 consistently strengthens motion-text alignment and motion fidelity: R@1/R@3 improve from 0.505/0.795 to 0.551/0.838, FID decreases from 0.323 to 0.070, and MM Dist drops from 3.305 to 2.745. This confirms that guidance is important for masked bidirectional sampling, where the model must progressively select confident tokens under the textual condition. However, further increasing the scale to 5.0 no longer brings additional gains, slightly degrading R@1, R@3, FID, and MM Dist. {We therefore set $w=4.0$ as the default CFG scale, balancing text-motion alignment and motion quality.}

\noindent\textbf{Effect of semi-autoregressive sampling.}
Following Block Diffusion~\cite{arriola2025block}, Table~\ref{tab:ab_block} compares the standard parallel sampling of BiMoGen with semi-autoregressive sampling. In semi-autoregressive sampling, each block is denoised in parallel, while blocks are generated sequentially. The results show that parallel sampling consistently achieves the best performance across all metrics, suggesting that preserving global bidirectional refinement is more effective than imposing sequential block dependencies. Among the semi-autoregressive variants, smaller blocks perform better, whereas larger blocks progressively degrade both alignment and motion quality.

\subsection{User Study Protocol}\label{supp:user_study}
We further conduct a user study to assess the perceptual quality of generated motions. We sample 30 text descriptions from the HumanML3D test set and generate motions under the same prompts for all compared methods. We recruit 15 users and present the generated motions in random order. For the comparison with representative text-to-motion baselines, users are asked to select the best result among MotionLCM, MotionGPT, and BiMoGen in terms of motion-text alignment, physical fidelity, and coherence/fluency. For the SC ablation, users compare BiMoGen with and without SC, with an additional ``Same'' option when no clear preference is observed.

\section{Sampling with Self-Correction}\label{supp:algo}
Algorithm~\ref{alg:sampling} gives the pseudo-code of sampling with self-correction. Given condition tokens $x^\mathrm{cond}$, BiMoGen iteratively refines a fully masked target: at each step, the model predicts clean target tokens under classifier-free guidance, and $\operatorname{Remask}(\cdot)$ then remasks the lowest-confidence predictions following LLaDA~\cite{llada}. Remasking is restricted to positions masked at the start of the current step, so tokens unmasked earlier are preserved. Here, $x$, $s$, $t$, and $T$ denote the current state, prediction confidences, current step, and total number of steps. At steps in $R$, self-correction completes the current target, then revises the visible tokens, leaving the remaining masked tokens to later steps.

\begin{algorithm}[!th!b]

\caption{Iterative Masked Sampling with Self-Correction}
\label{alg:sampling}

\begin{algorithmic}[1]
\Require Condition $x^\mathrm{cond}$, target length $L$, model $p_\theta$, steps $T$, CFG scale $w$, correction steps $\mathcal{R}$
\Ensure Generated target tokens $\widehat{x}$
\State $x \leftarrow[M]^L$
\For{$t=1$ to $T$}
    \State $P^\mathrm{cond}\leftarrow p_\theta(\cdot\mid x^\mathrm{cond},x)$, \quad $P^\mathrm{ucond}\leftarrow p_\theta(\cdot\mid \emptyset,x)$
    \State $P\leftarrow w P^\mathrm{cond}  + (1-w) P^\mathrm{ucond}$ 
    \State $y_i\leftarrow\operatorname*{arg\,max}_{v}P_i(v)$, \quad $s_i\leftarrow\max_v P_i(v)$ for $i$ with $x_i=[M]$
    \If{$t\in \mathcal{R}$}
        \State $\widetilde{x}\leftarrow x$
        \State $\widetilde{x}_i\leftarrow y_i$ for $i$ with $x_i=[M]$ \Comment{temporary complete target}
        \State $Q\leftarrow p_\theta(\cdot\mid x^\mathrm{cond},\widetilde{x})$
        \State $x_i\leftarrow\operatorname*{arg\,max}_{v}Q_i(v)$ for $i$ with $x_i\neq[M]$ \Comment{revise committed tokens}
    \EndIf
    \State $\bar{x}_i\leftarrow y_i$ for $i$ with $x_i=[M]$, and $\bar{x}_i\leftarrow x_i$ otherwise

    \State $x\leftarrow{\operatorname{Remask}(\bar{x};x,s,t,T)}$ \Comment{remask low-confidence predictions}

\EndFor
\State $\widehat{x}\leftarrow x$
\State \Return $\widehat{x}$
\end{algorithmic}
\end{algorithm}
\vspace{-10pt}
\section{Broader Impact}\label{supp:impact}

\noindent \textbf{Broader impact.}
This work may benefit motion content creation, animation, virtual agents, and motion retrieval through bidirectional language-motion generation. Since the model produces abstract motions rather than identifiable visual appearances, its direct risk is limited. Still, generated motions may be used in downstream synthetic avatar or animation systems, so responsible use, consent for real motion data, and disclosure of synthetic content should be considered in deployment.




\newpage

\section*{NeurIPS Paper Checklist}

\begin{enumerate}

\item {\bf Claims}
    \item[] Question: Do the main claims made in the abstract and introduction accurately reflect the paper's contributions and scope?
    \item[] Answer: \answerYes{} 
    \item[] Justification: The abstract and introduction state the main contributions, including the unified masked diffusion framework, decoupled training, and self-correction. The claims are supported by the experiments in Sec.~4 on HumanML3D and KIT-ML.


\item {\bf Limitations}
    \item[] Question: Does the paper discuss the limitations of the work performed by the authors?
    \item[] Answer: \answerYes{} 
    \item[] Justification: We discuss the main limitations in Sec.~\ref{limi}, including the dependence on VQ-VAE tokenization and the lack of explicit fine-grained body-part control.


\item {\bf Theory assumptions and proofs}
    \item[] Question: For each theoretical result, does the paper provide the full set of assumptions and a complete (and correct) proof?
    \item[] Answer: \answerNA{} 
    \item[] Justification: The paper does not present formal theoretical results, theorems, or proofs.


    \item {\bf Experimental result reproducibility}
    \item[] Question: Does the paper fully disclose all the information needed to reproduce the main experimental results of the paper to the extent that it affects the main claims and/or conclusions of the paper (regardless of whether the code and data are provided or not)?
    \item[] Answer: \answerYes{} 
    \item[] Justification: Sec.~\ref{sec4} reports the datasets, evaluation protocol, model configuration, training schedule, optimizer, sampling settings, and compute device used for the main experiments.


\item {\bf Open access to data and code}
    \item[] Question: Does the paper provide open access to the data and code, with sufficient instructions to faithfully reproduce the main experimental results, as described in supplemental material?
    \item[] Answer: \answerNo{} 
    \item[] Justification: We do not currently provide full reproduction scripts.


\item {\bf Experimental setting/details}
    \item[] Question: Does the paper specify all the training and test details (e.g., data splits, hyperparameters, how they were chosen, type of optimizer) necessary to understand the results?
    \item[] Answer: \answerYes{} 
    \item[] Justification: Sec.~\ref{sec4} specifies the datasets, metrics, model architecture, optimizer, learning rates, training epochs, task mixture ratio, condition dropout, and inference settings.


\item {\bf Experiment statistical significance}
    \item[] Question: Does the paper report error bars suitably and correctly defined or other appropriate information about the statistical significance of the experiments?
    \item[] Answer: \answerYes{} 
    \item[] Justification: For the T2M and M2T experiments, we follow the standard evaluation protocol and report averaged results over 20 evaluation runs, rather than single-run point estimates.


\item {\bf Experiments compute resources}
    \item[] Question: For each experiment, does the paper provide sufficient information on the computer resources (type of compute workers, memory, time of execution) needed to reproduce the experiments?
    \item[] Answer: \answerNo{} 
    \item[] Justification: Table~\ref{tab:ab_inference} reports parameter counts, FLOPs, and latency for the main model variants, and Sec.~\ref{sec4} states that experiments are run on a single NVIDIA H100.

    
\item {\bf Code of ethics}
    \item[] Question: Does the research conducted in the paper conform, in every respect, with the NeurIPS Code of Ethics \url{https://neurips.cc/public/EthicsGuidelines}?
    \item[] Answer: \answerYes{} 
    \item[] Justification: The work uses standard public motion-language benchmarks and does not involve new human-subject data collection. We have reviewed the NeurIPS Code of Ethics and believe the research conforms to it.


\item {\bf Broader impacts}
    \item[] Question: Does the paper discuss both potential positive societal impacts and negative societal impacts of the work performed?
    \item[] Answer: \answerYes{} 
    \item[] Justification: The paper discusses positive impacts on motion generation, animation, virtual agents, and assistive content creation, and also notes risks related to synthetic content misuse, dataset bias, and safety critical deployment.
    
\item {\bf Safeguards}
    \item[] Question: Does the paper describe safeguards that have been put in place for responsible release of data or models that have a high risk for misuse (e.g., pre-trained language models, image generators, or scraped datasets)?
    \item[] Answer: \answerNA{} 
    \item[] Justification: The paper does not release high-risk scraped datasets, pretrained language models, image generators, or other assets requiring special release safeguards.


\item {\bf Licenses for existing assets}
    \item[] Question: Are the creators or original owners of assets (e.g., code, data, models), used in the paper, properly credited and are the license and terms of use explicitly mentioned and properly respected?
    \item[] Answer: \answerNo{} 
    \item[] Justification: The paper cites the existing datasets, models, and evaluation assets used, but the current draft does not explicitly list their licenses, versions, or terms of use.


\item {\bf New assets}
    \item[] Question: Are new assets introduced in the paper well documented and is the documentation provided alongside the assets?
    \item[] Answer: \answerNA{} 
    \item[] Justification: The paper does not introduce or release new datasets, benchmarks, or model assets.


\item {\bf Crowdsourcing and research with human subjects}
    \item[] Question: For crowdsourcing experiments and research with human subjects, does the paper include the full text of instructions given to participants and screenshots, if applicable, as well as details about compensation (if any)? 
    \item[] Answer: \answerNo{} 
    \item[] Justification: Appendix~\ref{supp:user_study} summarizes the user study protocol. Participants received no monetary compensation. Full participant instructions are not included.


\item {\bf Institutional review board (IRB) approvals or equivalent for research with human subjects}
    \item[] Question: Does the paper describe potential risks incurred by study participants, whether such risks were disclosed to the subjects, and whether Institutional Review Board (IRB) approvals (or an equivalent approval/review based on the requirements of your country or institution) were obtained?
    \item[] Answer: \answerNA{} 
    \item[] Justification: The work does not involve new crowdsourcing or human-subject studies, so IRB or equivalent approval is not applicable.


\item {\bf Declaration of LLM usage}
    \item[] Question: Does the paper describe the usage of LLMs if it is an important, original, or non-standard component of the core methods in this research? Note that if the LLM is used only for writing, editing, or formatting purposes and does \emph{not} impact the core methodology, scientific rigor, or originality of the research, declaration is not required.
    \item[] Answer: \answerNA{} 
    \item[] Justification: LLMs are not used as an important, original, or non-standard component of the core method, nor are they used to generate data, labels, or evaluation results.

\end{enumerate}

\end{document}